\pdfoutput=1

\documentclass[11pt]{article}

\usepackage[]{acl}

\usepackage{times}
\usepackage{latexsym}

\usepackage[T1]{fontenc}

\usepackage[utf8]{inputenc}

\usepackage{microtype}

\usepackage{inconsolata}

\usepackage{times}
\usepackage{latexsym}
\usepackage{booktabs}
\usepackage{graphicx}
\usepackage{multirow}
\usepackage{times}
\usepackage{amsfonts,amsmath,amssymb,bm}
\usepackage{graphicx}
\usepackage{xcolor}
\usepackage{setspace}
\usepackage{pdflscape}
\usepackage{pdfpages}
\usepackage{enumitem}
\usepackage{rotating}
\usepackage{caption}
\usepackage{subcaption}

\newcommand{\asahi}[1]{\textcolor{black}{#1}}

\title{A \textsc{RelEntLess} Benchmark\\ for Modelling Graded Relations between Named Entities}

\author{Asahi Ushio \and Jose Camacho Collados \and Steven Schockaert\\
  Cardiff NLP, Cardiff University, UK\\ 
  \texttt{\{UshioA,CamachoColladosJ,SchockaertS1\}@cardiff.ac.uk} \\}

\begin{document}
\maketitle

\begin{abstract}
Relations such as ``is influenced by'', ``is known for'' or ``is a competitor of'' are inherently graded: we can rank entity pairs based on how well they satisfy these relations, but it is hard to draw a line between those pairs that satisfy them and those that do not. Such graded relations play a central role in many applications, yet they are typically not covered by existing Knowledge Graphs. In this paper, we consider the possibility of using Large Language Models (LLMs) to fill this gap. To this end, we introduce a new benchmark, in which entity pairs have to be ranked according to how much they satisfy a given graded relation. The task is formulated as a few-shot ranking problem, where models only have access to a description of the relation and five prototypical instances. We use the proposed benchmark to evaluate state-of-the-art relation embedding strategies as well as several
publicly available LLMs and closed conversational models such as GPT-4. 
We find that smaller language models struggle to outperform a naive baseline.
Overall, the best results are obtained with the 11B parameter Flan-T5 model and the 13B parameter OPT model, where further increasing the model size does not seem to be beneficial. For all models, a clear gap with human performance remains.
\end{abstract}

\section{Introduction}
Language Models (LMs) capture an abundance of factual and commonsense knowledge about the world \cite{petroni-etal-2019-language,roberts-etal-2020-much,heinzerling-inui-2021-language,west-etal-2022-symbolic,hao2022bertnet,cohen-etal-2023-crawling}. Given two entities, Large Language Models (LLMs) can straightforwardly be used to obtain a description of how these entities are related, although with some caveats for less popular entities \cite{mallen2022not}. However, relations are often a matter of degree \cite{rosch1975cognitive,turney-2006-similarity,vulic-etal-2017-hyperlex}. For instance, suppose we are interested in modelling whether one entity has been \emph{influenced by} another one. While we could argue that most contemporary pop music has been influenced by the Beatles, clearly there are some bands that have been influenced more directly than others. 
Graded relations such as \emph{influenced by}, \emph{competitor of} or \emph{similar to} are typically not found in traditional Knowledge Graphs (KGs), while they can nonetheless be of central importance to applications. For instance, in the context of financial NLP, we may need to know which companies are leaders and which are followers in a given field, who is competing with whom, and what strategic alliances exist. As another example, music recommendation systems often suggest artists based on the user's listening history, but these suggestions would be more helpful if the system could identify artists that have influenced or were influenced by artists the user already likes, as opposed to merely identifying similar artists. Studying how such relations can be modelled is thus clearly an important but under-explored research problem. 

The subjective nature of graded relations makes it difficult to include them in traditional KGs. Moreover, for many of these relations, it would simply not be feasible to list all the (graded) instances in a comprehensive way. Taking inspiration from existing work on extracting KGs from LLMs, we therefore ask the following question: \emph{are current LLMs capable of modelling graded relations between named entities in a meaningful way?}
The task of modelling graded relations offers a number of unique challenges for LLMs. First, since this is essentially a ranking task, designing suitable prompts is not straightforward. Second, the task requires making very fine-grained distinctions. For instance, while we can say that \emph{Microsoft is known for Windows} and \emph{Apple is known for MacOS}, the former statement represents a more prototypical instance of the \emph{known for} relation, as Apple is perhaps best known for its hardware products (e.g.\ iPhone). It is currently unclear to what extent LLMs are able to capture such subtle differences. Finally, modelling graded relations requires comparing entities of different types. For instance, the \emph{known for} relation has instances such as (\emph{Microsoft},\emph{Windows}), (\emph{the Beatles}, \emph{Hey Jude}) and even (\emph{France},\emph{wine}). Comparing instances of such a diverse nature poses a particular challenge, as such comparisons are almost never expressed in text. 

In this paper, we introduce \textsc{RelEntLess}\footnote{The name \textsc{RelEntLess} refers to \underline{Rel}ations between \underline{Ent}ities, where \underline{Less} refers to the idea of ordering. The dataset is available at \url{https://huggingface.co/datasets/cardiffnlp/relentless}.}, a new dataset aimed at furthering the study of graded relations between named entities.  Our dataset covers five common graded relations: competitor/rival of, friend/ally of, influenced by, known for, and similar to. We evaluate the ability of LLMs to rank entity pairs according to how much they satisfy these relations, given a description of the relation and five prototypical examples. Analysing the performance of several recent LLMs \cite{https://doi.org/10.48550/arxiv.2210.11416,iyer2022opt}, including GPT-4 \cite{openai2023gpt}, we find the best models to achieve a Spearman rank correlation of around 0.6. This shows that recent LLMs capture fine-grained relational knowledge to a meaningful extent, while at the same time still leaving a significant gap with human performance. For the open-source LLMs, we find that while the largest models achieve strong results, smaller models fail to outperform a naive baseline based on fastText vectors \cite{bojanowski-etal-2017-enriching}. GPT-3 performs well, albeit slightly below the best variants of Flan-T5 and OPT. Finally, we found ChatGPT and GPT-4 hard to use for this task, since the OpenAI API\footnote{\url{https://openai.com/blog/openai-api}} does not allow computing perplexity scores. As a result, we were not able to outperform GPT-3 with these models.

\section{Related Work}
\paragraph{Benchmarks for Graded Relations}
\textsc{RelEntLess} was inspired by the SemEval 2012 Task 2 dataset on modelling relational similarity \cite{jurgens-etal-2012-semeval}, which we will refer to as \emph{RelSim}. RelSim covers 79 fine-grained relations, which are organised into 10 categories, such as \emph{part-whole} (e.g.\ car:engine), \emph{attribute} (e.g.\ beggar:poor) and \emph{cause-purpose} (enigma:puzzlement). For each of the fine-grained relations, a ranking of concept pairs is provided, which reflects how prototypical these pairs are as instances of the relation. However, RelSim only considers concepts, whereas our focus is on named entities. To the best of our knowledge, the problem of modelling relational similarity between named entities has not yet been considered.

HyperLex \cite{vulic-etal-2017-hyperlex} is focused on modelling hypernymy as a graded relation. It involves ranking concept pairs according to how prototypical they are of the hypernymy relation. As for RelSim, named entities were explicitly excluded from this dataset. More broadly, word similarity benchmarks also follow the format of ranking concept pairs according to the degree to which a graded relation is satisfied, i.e.\ similarity. 

Benchmarks with analogy questions \cite{DBLP:conf/ranlp/TurneyLBS03,ushio-etal-2021-bert,chen-etal-2022-e} also relate to the problem of modelling graded relations. These benchmarks typically follow a multiple-choice format, where one word pair is given (e.g.\ eye:seeing), and the system has to predict which among a given set of candidate answer pairs is most analogous to the query pair (e.g.\ ear:hearing). Most existing benchmarks again focus on concepts. Moreover, where named entities are involved, the task degenerates to predicting whether two entity pairs have the same relation, i.e.\ the problem of measuring degrees of relatedness is not considered for named entities. 

\smallskip
\paragraph{Language Models as Knowledge Bases}
The idea of using language models as knowledge bases was popularised by \citet{petroni-etal-2019-language}, and has gained considerable further traction with the advent of LLMs. For instance, several authors have proposed strategies for extracting knowledge graphs from LLMs \cite{west-etal-2022-symbolic,hao2022bertnet,cohen-etal-2023-crawling}. While the idea of modelling graded relations has not been considered, \citet{hao2022bertnet} focused on relations that are not covered by traditional knowledge graphs, such as ``is capable of but not good at''. Similarly, our motivation for studying graded relations between named entities is also to complement what is captured by KGs.

\begin{table*}[!t]
\centering
\footnotesize
\begin{tabular}{lccp{150pt}p{150pt}}
\toprule
         \textbf{Relation Type} &  \textbf{Val}  &  \textbf{Test} &    \multicolumn{1}{c}{\textbf{Prototypical examples}} &   \multicolumn{1}{c}{\textbf{Middle rank examples}} \\
\midrule
\multirow{3}{*}{competitor/rival of} & \multirow{3}{*}{20} & \multirow{3}{*}{84}      &  Dell : HP, Sprite : 7 Up, Israel : Palestine, Liverpool FC : Manchester United, Microsoft Teams : Slack        &  Macallan : Suntory, Marvel Comics : D.C. Comics, Borussia Dortmund : PSG, UK : France, Doctor Who : Game of Thrones \\   
\midrule
\multirow{3}{*}{friend/ally of} & \multirow{3}{*}{20} &   \multirow{3}{*}{88} &     Australia : New Zealand, Aznar : Bush, Extinction Rebellion : Greta Thunberg, Elsa : Anna, CIA : MI6     & Kylo Ren : Rey, UK : Commonwealth, Darth Vader : Emperor Palpatine, The Beatles : Queen, Mark Drakeford : Rishi Sunak\\
\midrule
\multirow{3}{*}{influenced by} & \multirow{3}{*}{20} &    \multirow{3}{*}{90}        &  Europe : European Union, Plato : Socrates, Ethereum : Bitcoin, Messi : Maradona, Impressionism : Edouard Manet & Mike Tyson : Muhammad Ali, US : NASA, Acer : Asus, Vincent van Gogh : Bipolar disorder, Conservative Party : Labour Party\\
\midrule
\multirow{3}{*}{known for} & \multirow{3}{*}{20} &    \multirow{3}{*}{105}        &  Russell Crowe : Gladiator, Cadbury : chocolate, Paris : Eiffel Tower, Leonardo Da Vinci : Mona Lisa, Apple : iPhone & New Zealand : sheep, Le Corbusier : purism art, Sean Connery : Finding Forrester, Qualcomm : smartphones, Nikola Tesla : robotics\\
\midrule
\multirow{3}{*}{similar to} & \multirow{3}{*}{20} &    \multirow{3}{*}{89} &               Coca-Cola : Pepsi, Ligue 1 : Bundesliga, Australia : New Zealand, The Avengers : The Justice League, Tesco : Sainsburys & NATO : United Nations, Iraq : Iran, cement : concrete, Cornwall : Brittany, Adele : Ed Sheeran \\ 
\bottomrule
\end{tabular}
\caption{Overview of the considered relations, showing the numbers of entity pairs in the validation and test sets, the five prototypical training examples, and five examples from the middle of the ranking of the entity pairs in the validation set.}
\label{tab:data-statistics}
\end{table*}

\begin{table}[!t]
\centering
\footnotesize
\begin{tabular}{l@{\hspace{5pt}}p{195pt}}
\toprule
5:& This is clearly a positive example, and I would expect everyone to agree with this view.\\[0.5em]
4:& I consider this to be a positive example, but I would not be surprised if some knowledgeable people consider this word pair to be borderline.\\[0.5em]
3:& I consider this to be a borderline case: I find it hard to decide whether this is a positive or a negative example. \\[0.5em]
2:& I consider this to be a negative example, but I would not be surprised if some knowledgeable people consider this word pair to be borderline. \\[0.5em]
1:& This is clearly a negative example, and I would expect everyone to agree with this view.\\
\bottomrule
\end{tabular}
\caption{Rating scale for the 2nd annotation phase.\label{tabScale} }
\end{table}

\section{Dataset}
We consider the five relations which are shown in \autoref{tab:data-statistics}. These relations were chosen because of their graded character and because they can apply to a broad range of entities. We created a dataset with annotated entity pairs for each of the relations in three phases. We recruited a diverse annotation team in terms of age, gender, ethnicity and nationality; however, all annotators come from an academic setting: four undergraduate students, one PhD student and two faculty members. The students were recruited through an internal student employment service and were offered a remuneration of around \$20 per hour. The total annotation effort was about 160 hours. The annotation process was split into three phases.

\paragraph{First phase} In the first phase, the annotators were asked to provide 15 entity pairs for each of the five relations. Specifically, the aim was to provide 5 prototypical examples (i.e.\ entity pairs that clearly satisfy the relationship), 5 borderline positive pairs, which only satisfy the relationship to some extent, and 5 borderline negative pairs, which do not satisfy the intended relationship but are nonetheless related in a similar way. After removing duplicates, this resulted in an average of 114 entity pairs for each relation, and 573 pairs in total. We augmented these entity pairs with the same number of randomly chosen entity pairs as the annotated pairs in each relation type. The entities for these random pairs were selected from the 50,000 most popular Wikidata entities, in terms of the number of page views of the associated Wikipedia article. 

\paragraph{Second phase} In the second phase, each annotator scored all the entity pairs that were provided in phase 1, using the 5-point scale shown in \autoref{tabScale}. For this phase, annotators were encouraged to consult web sources (e.g.\ search engines such as Google) for a limited time in order to familiarize themselves with the considered entities, if needed. This was the most time-consuming annotation phase, taking almost 10 hours on average per annotator to complete.

\paragraph{Third phase} The third and final phase was aimed at resolving disagreements between the annotations from the second phase. Specifically, for each entity pair where there was a difference of 3 points between the highest and the lowest score, the annotator(s) with a diverging view were asked to check their previous annotation, and to either update their score or to provide a justification. A total of 255 unique entity pairs were checked in this way (310 scores were checked in total). We subsequently verified the justifications that were provided. In 13 cases, the justifications suggested that the other annotators might have missed a salient point. For these cases, the annotators with the opposite view were asked to re-check their previous annotation. The final ranking for each relation was obtained by averaging the scores of the 7 annotators.

\begin{table}[!t]
\centering
\footnotesize
\begin{tabular}{c@{\hspace{7pt}}c@{\hspace{7pt}}c@{\hspace{7pt}}c@{\hspace{7pt}}c@{\hspace{7pt}}c@{\hspace{7pt}}c@{\hspace{7pt}}c@{\hspace{7pt}}c}
\toprule
{} &    A &    B &    C &    D &    E &    F &    G &  Others \\
\midrule
A   &  100 &   62 &   81 &   71 &   75 &   75 &   75 &      84 \\
B   &   62 &  100 &   61 &   57 &   62 &   57 &   60 &      66 \\
C   &   81 &   61 &  100 &   73 &   72 &   74 &   75 &      84 \\
D   &   71 &   57 &   73 &  100 &   67 &   67 &   70 &      77 \\
E   &   75 &   62 &   72 &   67 &  100 &   70 &   72 &      77 \\
F   &   75 &   57 &   74 &   67 &   70 &  100 &   69 &      76 \\
G   &   75 &   60 &   75 &   70 &   72 &   69 &  100 &      79 \\ \midrule
AVG &   77 &   66 &   77 &   72 &   74 &   73 &   74 &      77 \\
\bottomrule
\end{tabular}
\caption{Spearman correlation (\%) between each pair of annotators (A,...,G), and between each annotator and the average score provided by the other six averaged over all the five relation types after the 3rd and final quality enhancement annotation round.}
\label{tab:annotator-agreement}
\end{table}

\autoref{tab:annotator-agreement} summarises the agreement between the annotators in terms of Spearman's rank correlation.\footnote{In \autoref{app:iaa}, we include the breakdown of the annotator agreement scores per relation type.} The table shows the correlation between the individual annotators, as well as the correlation between each annotator and the average of the scores from the six other annotators. The reconciliation step improved the average agreement over all the annotators from 70 to 77.\footnote{Details about the agreement before the reconciliation step can be found in the appendix.}

We split the annotated entity pairs as follows. First, we selected a small training set consisting of five prototypical pairs for each relation. This training set could be used, for instance, for few-shot prompting strategies. The entity pairs were selected (i) to be among the top-ranked entity pairs and (ii) to be sufficiently diverse (i.e.\ including entities of different types). Next, for each relation, we randomly selected 20 of the remaining entity pairs to be used as a validation set.\footnote{This validation set was not used in our main experiments, but it was considered in the few-shot analysis (see \autoref{secZeroFewShot}). However, we release the full validation set so it can be used for further testing and experimentation without the risk of overfitting on the test set} The remaining entity pairs constitute the test set. \autoref{tab:data-statistics} shows the prototypical entity pairs that were selected for each relation, as well as five examples of entity pairs from the validation set. The latter were selected from the middle of the ranking, typically with an average score of 3 to 4. We use the Spearman rank correlation between the predicted ranking and the ground truth ranking as the evaluation metric.\footnote{The final annotated dataset, along with the guidelines provided to annotators in each phase, are available in the supplementary material.}

\section{Baselines}

\paragraph{Human Performance}
As a proxy for human performance, we report the average Spearman rank correlation between each annotator and the average of the other annotators, referred to as \emph{Human Upperbound}. Please note that this upperbound is computed based on the test set, and thus slightly differs from the average agreement in \autoref{tab:annotator-agreement}. 
Furthermore, note that we only estimate human performance to provide a reference for interpreting the results. Doing this accurately is challenging. For instance, we can already see large differences in agreement across the different annotators, suggesting that the best annotators would perform much better than what is suggested by the given upperbound. Conversely, one may also argue that because of the reconciliation step in the third phrase, we are overestimating human performance.

\subsection{Embedding Models}

\paragraph{Word Embedding.} First, we consider the fastText \cite{bojanowski-etal-2017-enriching} embeddings that were trained on Common Crawl with subword information\footnote{\url{https://fasttext.cc/}}. Inspired by the tradition of modelling word analogies using vector differences \cite{mikolov-etal-2013-linguistic}, we represent each entity pair by subtracting the fastText embedding of the first entity from the embedding of the second entity. We refer to the resulting vector as the fastText relation embedding. For a given relation, we score an entity pair by taking the maximum cosine similarity between its fastText relation embedding and the embedding of the five prototypical examples.\footnote{Empirically, we confirmed that indeed using the maximum leads to better results overall.} We use the maximum, rather than e.g.\ the average, due to the diverse nature of these prototypical examples. We refer this approach as fastText\textsubscript{pair}.

As a naive baseline, we also consider a variant in which an entity pair is scored by taking the cosine similarity between the word embeddings of the two entities. Note that this baseline ignores both the description of the relation and the prototypical examples. It is based on the idea that prototypical pairs often involve closely related entities. We refer to this approach as fastText\textsubscript{word}.

\paragraph{RelBERT.} RelBERT \cite{ushio-etal-2021-distilling} is a RoBERTa model that was fine-tuned to encode word pairs such that analogous word pairs are represented by similar vectors. We use RelBERT models that were initialised from RoBERTa\textsubscript{BASE}\footnote{\url{https://huggingface.co/relbert/relbert-roberta-base}} and from RoBERTa\textsubscript{LARGE}\footnote{\url{https://huggingface.co/relbert/relbert-roberta-large}}. For a given relation, we score each entity pair as the maximum cosine similarity between its RelBERT encoding and the RelBERT encoding of the five prototypical examples. 

\subsection{Language Models}
\label{sec:langmodels}
To score entity pairs using LMs, we create a prompt from the description of the relation and the five prototypical examples. The score of the entity pair then corresponds to the perplexity of the prompt.
We consider two prompt templates: a binary question answering (QA) template similar to the instructions provided to Flan-T5 for the task \cite{longpre2023flan}, and a targeted list completion template (LC).
\asahi{Writing the five prototypical examples as $[A_i, B_i]_{i=1\dots5}$ and the target entity pair as $[C,D]$, the QA template has the following form:
\begin{quote}
Answer the question by yes or no. We know that $[A_1, B_1],\dots,[A_5, B_5]$ are examples of \texttt{<desc>}. Are $[C, D]$ \texttt{<desc>} as well? \\
Yes
\end{quote}
The LC template has the following form:
\begin{quote}
Complete the following list with examples of \texttt{<desc>} \\
$[A_1, B_1]$ \\
: \\
$[A_5, B_5]$ \\
$[C, D]$ 
\end{quote}
In both templates, \texttt{<desc>} is the description of the relation, as follows:
\begin{itemize}
\setlength\itemsep{0.2em}
\item \emph{Rival:} entities that are competitors or rivals
\item \emph{Ally:} entities that are friends or allies
\item \emph{Inf:} what has influenced different entities
\item \emph{Know:} what entities are known for
\item \emph{Sim:} entities that are similar
\end{itemize}}

We use the following LMs: OPT \cite{zhang2022opt}, OPT-IML \cite{iyer2022opt}, T5 \cite{2020t5}, Flan-T5 \cite{https://doi.org/10.48550/arxiv.2210.11416}, and Flan-UL2 \cite{tay2023ul2}, where the model weights are obtained via HuggingFace \cite{wolf-etal-2020-transformers}\footnote{A complete list of the models on huggingface we used can be found in \autoref{app:hf-name}.}. We also use GPT-3 \cite{GPT3}, which is a private model and subject to be changed every six months; we use \texttt{davinci}, which is the most powerful GPT-3 model available via the OpenAI API \footnote{\url{https://openai.com}}\footnote{All the OpenAI models are from the checkpoint that was live during May 2023.}.
We compute the perplexity over the whole input text for OPT, OPT-IML and GPT-3, while we use the last line of the input text (i.e., ``Yes'' for the QA template and $[C, D]$ for the LC template) to compute the perplexity on the decoder for T5, Flan-T5, and Flan-UL2.

We test two conversational LMs: ChatGPT (or \texttt{gpt-3.5-turbo}) and GPT-4 (\texttt{gpt-4}). These models are only available through the OpenAI API. Unfortunately, for these models, the API does not allow us to obtain the log-likelihood of each token. Therefore, we instead use a prompt which asks to sort the list of entity pairs directly.
\asahi{Writing the list of target word pairs as $[C_i,D_i]_{i=1\dots n}$, our prompt has the following form: 
\begin{quote}
Consider the following reference list of \texttt{<desc>}: \\
$[A_1, B_1]$ \\
:\\
$[A_5, B_5]$ \\
Now sort the entity pairs from the following list based on the extent to which they also represent \texttt{<desc>} in descending order. Do not include the pairs from the reference list. The output should contain all the entity pairs from the following list and no duplicates: \\
$[C_1, D_1]$ \\
: \\
$[C_n, D_n]$
\end{quote}
These conversational models often omit entity pairs from the output, especially those with lower similarity to the reference pairs. To deal with this, we simply concatenate those removed pairs to the bottom of the sorted output list. 
}

\section{Results}

\begin{table*}[!t]
\centering
\footnotesize
\begin{tabular}{llllccrrrrrr}
\toprule
{} &&&& Inst-FT & Model Size &  Rival &  Ally &   Inf &  Know &   Sim &   Average \\
\midrule
\multicolumn{6}{l}{\textit{Human Upperbound}} &                 75.9 &            78.0 &           70.5 &       82.0 &        80.2 &     77.3 \\
\midrule 
\multicolumn{3}{l}{\multirow{4}{*}{\rotatebox{0}{Embedding}}}
&\multicolumn{1}{l}{fastText\textsubscript{word}} && - &                 25.0 &            10.0 &            7.0 &       24.0 &        20.0 &     17.0 \\
&&&\multicolumn{1}{l}{fastText\textsubscript{pair}} &&  - 
&                 28.0 &            12.0 &            3.0 &       20.0 &        21.0 &     17.0 \\
&&&\multicolumn{1}{l}{RelBERT\textsubscript{BASE}} && 110M
&                 58.0 &            15.0 &           30.0 &       24.0 &        28.0 &     31.0 \\
&&&\multicolumn{1}{l}{RelBERT\textsubscript{LARGE}} && 335M 
&                 64.0 &            20.0 &           20.0 &       44.0 &        53.0 &     40.0 \\
\midrule 
\multirow{52}{*}{\rotatebox{0}{LM}} 
&\multirow{24}{*}{\rotatebox{90}{\emph{LC template}}} 

&\multirow{11}{*}{{T5}} 
&T5\textsubscript{SMALL}         &&60M
&                 20.0 &            33.0 &           24.0 &       11.0 &        10.0 &     19.0 \\
&&&T5\textsubscript{BASE}          &&220M
&                 35.0 &            35.0 &           38.0 &       20.0 &        13.0 &     28.0 \\

&&&T5\textsubscript{LARGE}         &&770M
&                 29.0 &             8.0 &           26.0 &       11.0 &        22.0 &     19.0 \\

&&&T5\textsubscript{XL}            &&3B
&                 47.0 &            28.0 &           50.0 &       33.0 &        26.0 &     37.0 \\

&&&T5\textsubscript{XXL}           &&11B
&                 33.0 &             8.0 &           24.0 &       18.0 &        15.0 &     19.0 \\
\cmidrule{4-12}

&&&Flan-T5\textsubscript{SMALL}    &$\checkmark$&60M
&                 38.0 &            33.0 &           24.0 &       16.0 &         7.0 &     24.0 \\

&&&Flan-T5\textsubscript{BASE}     &$\checkmark$&220M
&                 36.0 &            31.0 &           28.0 &       17.0 &        -0.0 &     22.0 \\

&&&Flan-T5\textsubscript{LARGE}    &$\checkmark$&770M
&                 41.0 &            19.0 &           36.0 &       24.0 &        22.0 &     29.0 \\

&&&Flan-T5\textsubscript{XL}       &$\checkmark$&3B
&                 40.0 &            17.0 &           35.0 &       27.0 &        31.0 &     30.0 \\

&&&Flan-T5\textsubscript{XXL}      &$\checkmark$&11B
&                 61.0 &            32.0 &           47.0 &       44.0 &        40.0 &     45.0 \\
\cmidrule{4-12}

&& &Flan-UL2                        &$\checkmark$&20B
&                 60.0 &            28.0 &           49.0 &       53.0 &        37.0 &     45.0 \\
\cmidrule{3-12}

&&\multirow{12}{*}{{OPT}} 
&OPT\textsubscript{125M}         &&125M
&                 41.0 &            37.0 &           51.0 &       23.0 &        13.0 &     33.0 \\

&&&OPT\textsubscript{350M}         &&300M
&                 41.0 &            33.0 &           47.0 &       36.0 &        18.0 &     35.0 \\

&&&OPT\textsubscript{1.3B}         &&1.3B
&                 58.0 &            39.0 &           54.0 &       45.0 &        42.0 &     48.0 \\

&&&OPT\textsubscript{2.7B}         &&2.7B
&                 65.0 &            41.0 &           58.0 &       56.0 &        42.0 &     52.0 \\

&&&OPT\textsubscript{6.7B}         &&6.7B
&                 71.0 &            42.0 &           59.0 &       61.0 &        47.0 &     56.0 \\

&&&OPT\textsubscript{13B}          &&13B
&                 72.0 &            41.0 &           55.0 &       70.0 &        55.0 &     59.0 \\

&&&OPT\textsubscript{30B}          &&30B
&                 71.0 &            39.0 &           57.0 &       69.0 &        53.0 &     58.0 \\
\cmidrule{4-12}

&&&OPT-IML\textsubscript{1.3B}     &$\checkmark$&1.3B
&                 57.0 &            39.0 &           56.0 &       51.0 &        35.0 &     47.0 \\
&&&OPT-IML\textsubscript{30B}      &$\checkmark$&30B
&                 65.0 &            36.0 &           55.0 &       70.0 &        47.0 &     55.0 \\
&&&OPT-IML\textsubscript{MAX-1.3B} &$\checkmark$&1.3B
&                 55.0 &            37.0 &           57.0 &       49.0 &        33.0 &     46.0 \\
&&&OPT-IML\textsubscript{MAX-30B}  &$\checkmark$&30B
&                 62.0 &            36.0 &           57.0 &       67.0 &        46.0 &     53.0 \\

\cmidrule{3-12}
&&GPT&GPT-3\textsubscript{davinci}*    & &-
&                 72.0 &            39.0 &           \textbf{64.0} &       \textbf{73.0} &        47.0 &     59.0 \\
\cmidrule{2-12}

&\multirow{24}{*}{\rotatebox{90}{\emph{QA template}}} 
&\multirow{11}{*}{{T5}} &T5\textsubscript{SMALL}         &&60M
&                 10.0 &           -13.0 &           17.0 &       -6.0 &         8.0 &      3.0 \\

&&&T5\textsubscript{BASE}          &&220M
&                 15.0 &            -7.0 &            6.0 &      -12.0 &        14.0 &      3.0 \\

&&&T5\textsubscript{LARGE}         &&770M
&                 -3.0 &             4.0 &          -12.0 &      -19.0 &        -1.0 &     -6.0 \\

&&&T5\textsubscript{XL}            &&3B
&                 -2.0 &            12.0 &           -8.0 &       17.0 &       -14.0 &      1.0 \\

&&&T5\textsubscript{XXL}           &&11B
&                  7.0 &             1.0 &           -1.0 &       11.0 &        -4.0 &      3.0 \\
\cmidrule{4-12}

&&&Flan-T5\textsubscript{SMALL}    &$\checkmark$&60M
&                 31.0 &            -0.0 &           21.0 &       -3.0 &         8.0 &     11.0 \\

&&&Flan-T5\textsubscript{BASE}     &$\checkmark$&220M
&                 41.0 &            28.0 &           46.0 &       17.0 &        22.0 &     31.0 \\

&&&Flan-T5\textsubscript{LARGE}    &$\checkmark$&770M
&                 67.0 &            39.0 &           24.0 &       49.0 &        56.0 &     47.0 \\

&&&Flan-T5\textsubscript{XL}       &$\checkmark$&3B
&                 75.0 &            44.0 &           44.0 &       61.0 &        63.0 &     57.0 \\

&&&Flan-T5\textsubscript{XXL}      &$\checkmark$&11B
&                 74.0 &            \textbf{56.0} &           44.0 &       70.0 &        66.0 &     \textbf{62.0} \\
\cmidrule{4-12}

&&&Flan-UL2                        &$\checkmark$&20B
&                 \textbf{79.0} &            51.0 &           47.0 &       67.0 &        57.0 &     60.0 \\
\cmidrule{3-12}
&&\multirow{12}{*}{{OPT}} 
&OPT\textsubscript{125M}         &&125M
&                 35.0 &            31.0 &           46.0 &       10.0 &         9.0 &     26.0 \\

&&&OPT\textsubscript{350M}         &&350M
&                 38.0 &            35.0 &           37.0 &       21.0 &        19.0 &     30.0 \\

&&&OPT\textsubscript{1.3B}         &&1.3B
&                 44.0 &            33.0 &           46.0 &       29.0 &        31.0 &     37.0 \\

&&&OPT\textsubscript{2.7B}         &&2.7B
&                 54.0 &            32.0 &           50.0 &       38.0 &        32.0 &     41.0 \\

&&&OPT\textsubscript{6.7B}         &&6.7B
&                 53.0 &            33.0 &           39.0 &       46.0 &        34.0 &     41.0 \\

&&&OPT\textsubscript{13B}          &&13B
&                 63.0 &            39.0 &           43.0 &       61.0 &        43.0 &     50.0 \\

&&&OPT\textsubscript{30B}          &&30B
&                 61.0 &            38.0 &           48.0 &       62.0 &        45.0 &     51.0 \\
\cmidrule{4-12}

&&&OPT-IML\textsubscript{1.3B}     &$\checkmark$&1.3B
&                 45.0 &            27.0 &           42.0 &       21.0 &        26.0 &     32.0 \\

&&&OPT-IML\textsubscript{30B}      &$\checkmark$&30B
&                 57.0 &            37.0 &           36.0 &       53.0 &        35.0 &     44.0 \\

&&&OPT-IML\textsubscript{MAX-1.3B} &$\checkmark$&1.3B
&                 42.0 &            25.0 &           38.0 &       16.0 &        29.0 &     30.0 \\

&&&OPT-IML\textsubscript{MAX-30B}  &$\checkmark$&30B
&                 58.0 &            36.0 &           39.0 &       43.0 &        42.0 &     43.0 \\

\cmidrule{3-12}
&&GPT&GPT-3\textsubscript{davinci}*    &&-
&                 67.0 &            35.0 &           50.0 &       61.0 &        35.0 &     50.0 \\

\midrule
\multicolumn{3}{l}{\multirow{2}{*}{\rotatebox{0}{Conv. LM}}}
&ChatGPT*    &&-&   -0.9 &  32.5 &  17.5 &  15.5 &  14.7 &  17.9 \\
&&&GPT-4*      &&-&  62.5 &  55.8 &  35.9 &  60.8 &   \textbf{69.3} &  56.9\\
\midrule
\multicolumn{5}{l}{LM Ensemble} & - & \textit{78.9} & \textit{50.1} & \textit{61.6} & \textit{75.5} & \textit{65.9} & \textit{66.4} \\
\bottomrule
\end{tabular}
\caption{Spearman's rank correlation (\%) on the test set. The LMs are grouped by the template (QA or LC), the model family, and instruction-fine-tuned or not. The best correlation in each relation type is highlighted by bold characters, \asahi{except for LM ensemble emphasized by italic}. 
Model size is measured as the number of parameters. Models marked with * are not openly available.
}
\label{tab:baselines}
\end{table*}

\autoref{tab:baselines} summarises the results. The best result is achieved by Flan-T5\textsubscript{XXL} with the QA template, which scores 62.0\%. In general, the performance of this model remains far below the performance upper bound suggested by the inter-annotator agreement (77\%). Surprisingly, however, for the \emph{rival of} relation, the human upper bound is outperformed by Flan-UL2. In contrast, the \emph{friend/ally of} relation appears to be particularly challenging.  Among the LM methods, the LC template generally leads to the best results, but not for Flan-T5 and Flan-UL2.  This is not entirely surprising given that Flan models have been fine-tuned using instructions similar to the QA template (see \autoref{sec:langmodels}). Beyond the encoder-decoder LMs, OPT\textsubscript{13B} and GPT-3\textsubscript{davinci} perform the best, even outperforming the instruction fine-tuned OPTs (OPT-IML and OPT-IML\textsubscript{MAX}). GPT-3\textsubscript{davinci} is the best model in the \emph{influenced by} and \emph{known for} relations. Although Flan-T5\textsubscript{XXL} and Flan-UL2 perform best on average, they perform poorly on the \emph{influenced by} relation, underperforming GPT-3\textsubscript{davinci} and OPT\textsubscript{13B} by a wide margin.  Among the embedding based models, fastText generally performs poorly. The performance of RelBERT\textsubscript{LARGE} is remarkably strong, considering that this is a small concept-based relation model that was not trained on relations between named entities. As far as the OpenAI conversational models are concerned, we can see that GPT-4 achieves the best result on the \emph{similar to} relation. The poor performance of ChatGPT suggests that the considered list ranking prompt may be hard to understand for this model, or that the task of ranking around 100 pairs may be too complicated. We also observed that ChatGPT tends to omit more pairs from its output than GPT-4 \asahi{(see \autoref{tab:chat-model} that shows the results and percentage of retrieved pairs of the conversational LMs.
)}.

\begin{table}[!t]
\centering
\footnotesize
\begin{tabular}{lcc}
\toprule
{} &       ChatGPT &         GPT-4 \\
\midrule
Rival &   -0.9 (0.0\%) &  62.5 (100.0\%) \\
Ally  &  42.5 (56.8\%) &  55.8 (100.0\%) \\
Inf   &  17.5 (91.1\%) &   35.9 (94.4\%) \\
Know  &  15.5 (86.7\%) &  60.8 (100.0\%) \\
Sim   &  14.7 (80.9\%) &   69.3 (98.9\%) \\ \midrule
AVG   &  17.9 (63.1\%) &   56.9 (98.7\%) \\ 
\bottomrule
\end{tabular}
\caption{Spearman's rank correlation (\%) on the test set for conversational LMs with the percentage of word pairs included in the output.}
\label{tab:chat-model}
\end{table}

\asahi{We also report the result of a simple model ensemble (denoted as LM ensemble on \autoref{tab:baselines}), where we choose the top-5 models regarding to the average accuracy (Flan-UL2 with QA template, Flan-T5\textsubscript{XXL} with QA template, OPT\textsubscript{13B} with LC template, OPT\textsubscript{30B} with LC template, and GPT-3\textsubscript{davinci} with LC template), and we use the averaged perplexity across these five models to compute the ranking. As can be seen in \autoref{tab:baselines}, this indeed leads to better results on average, although not consistently for all relations.}

\begin{figure}[!t]
    \centering
    \begin{subfigure}[b]{0.47\columnwidth}
        \centering
        \includegraphics[width=\columnwidth]{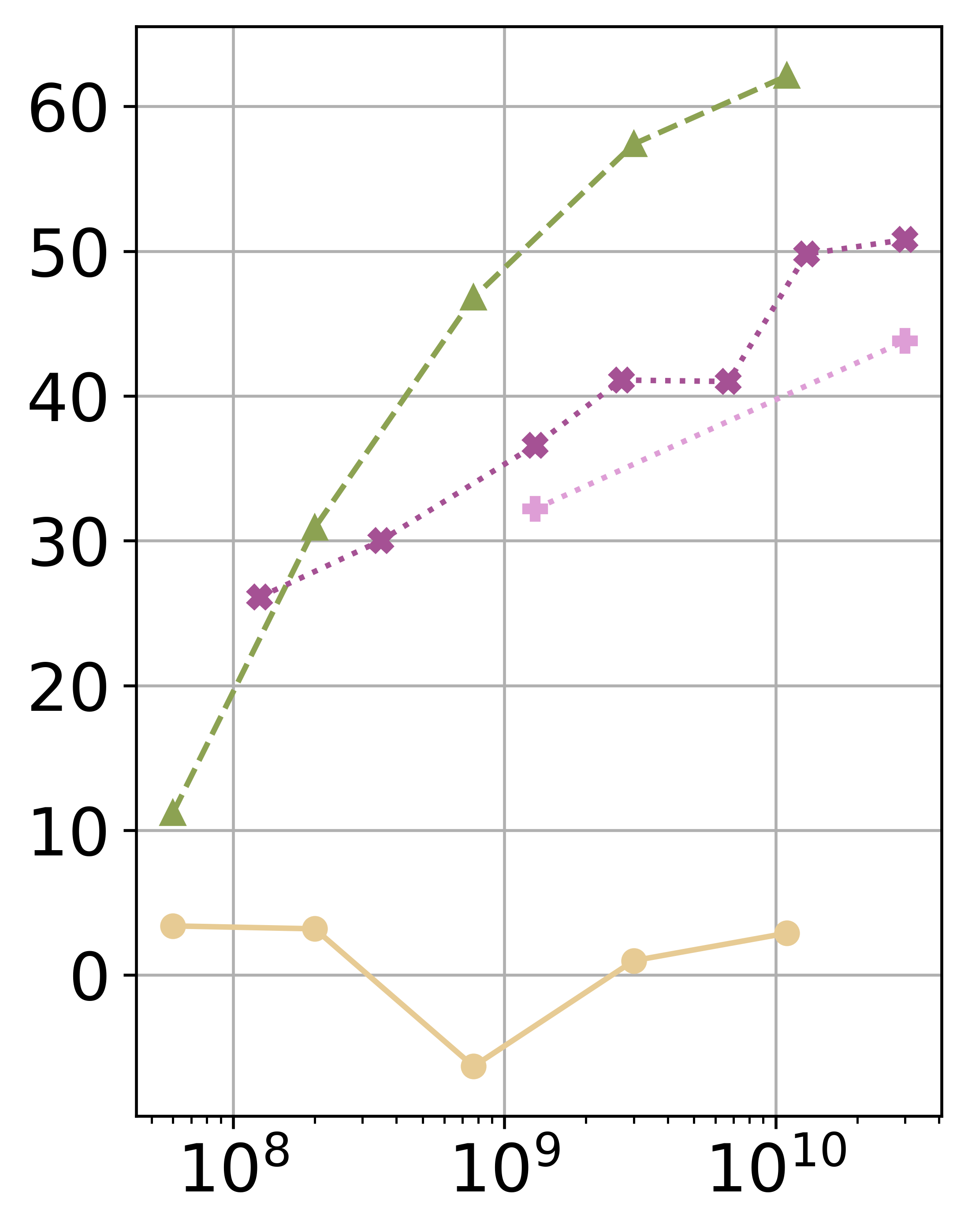}
        \caption{QA template}
    \end{subfigure}     
    \begin{subfigure}[b]{0.47\columnwidth}
        \centering
        \includegraphics[width=\columnwidth]{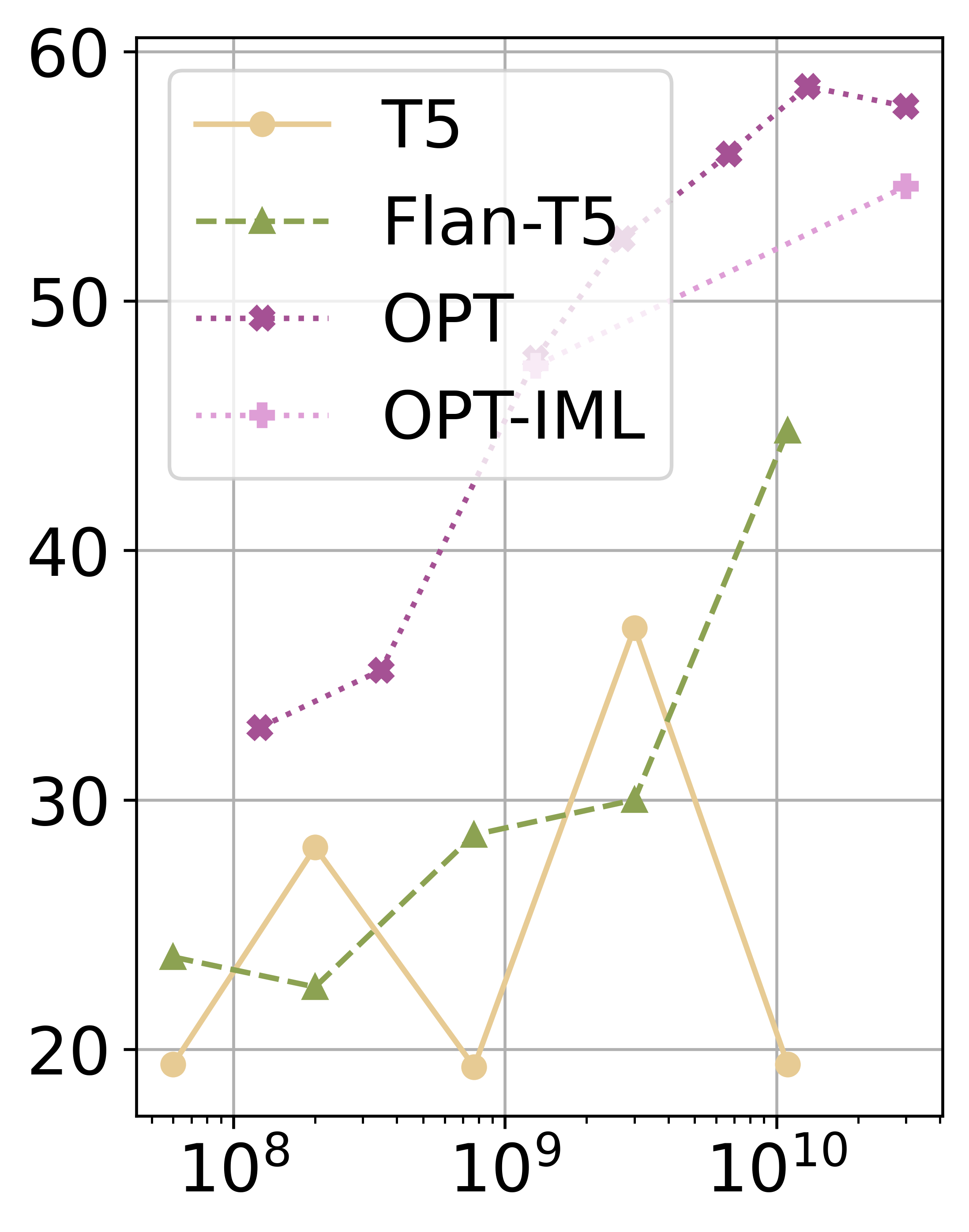}
        \caption{LC template}
    \end{subfigure}
    \caption{Average Spearman's rank correlation results among the five relation types along with the model size.}
    \label{fig:correlation-with-model-size-ave}
\end{figure}

\section{Analysis}
We now aim to gain a better understanding of the behaviour of LMs. First, we analyse the effect of model size (\autoref{secmodelsize}). Then, we experiment with different zero-shot and few-shot learning set-ups (\autoref{secZeroFewShot}), and we present a qualitative analysis of the predictions (\autoref{secQualitative}). For the latter two analyses, we focus on the best performing models for each LM family from the main experiment, using their optimal prompts: Flan-UL2, Flan-T5\textsubscript{XXL}, OPT\textsubscript{13B}, and GPT-3\textsubscript{davinci}.\footnote{Note that we omit Flan-UL2 from the model size analysis as there is only a single Flan-UL2 model.}

\subsection{Model Size}\label{secmodelsize}
In this section, we analyse the effect of model size. \autoref{fig:correlation-with-model-size-ave} visualises the performance of the different model families in function of model size. For Flan-T5, OPT, and OPT-IML we can see a strong correlation between performance and size. Nevertheless, the result of the largest OPT models suggests that a plateau in performance may have been reached at 13B. Moreover, for T5 we do not see an improvement in performance for larger models\footnote{In  \autoref{app:additional-result} we include a more detailed breakdown of the results of this model size experiment by relation type.}.

\subsection{Zero-shot/Few-shot Learning}\label{secZeroFewShot}
In the main experiments, for each relation, models had access to a description as well as five prototypical examples. To analyse the impact of these five examples, we now describe experiments in which only the description is provided (i.e.\ zero-shot) or where only 1 or 3 examples are given (few-shot). For the few-shot setting, we use the same QA and LC templates as in the main experiment. For the 3-shot experiments, we randomly choose 3 of the 5 examples, and similar for the 1-shot experiments. Since this introduces some randomness, we report results for three different samples.

The QA template for zero-shot/few-shot learning are:
\begin{quote}
Answer the question by yes or no. Are $[C, D]$ \texttt{<desc>}?\\
Yes
\end{quote}
while the zero-shot LC template has the following form:
\begin{quote}
Complete the following list with examples of \texttt{<desc>}?\\
$[C, D]$
\end{quote}

\begin{figure}[!t]
    \centering
    \begin{subfigure}[b]{0.47\columnwidth}
        \centering
        \includegraphics[width=\columnwidth]{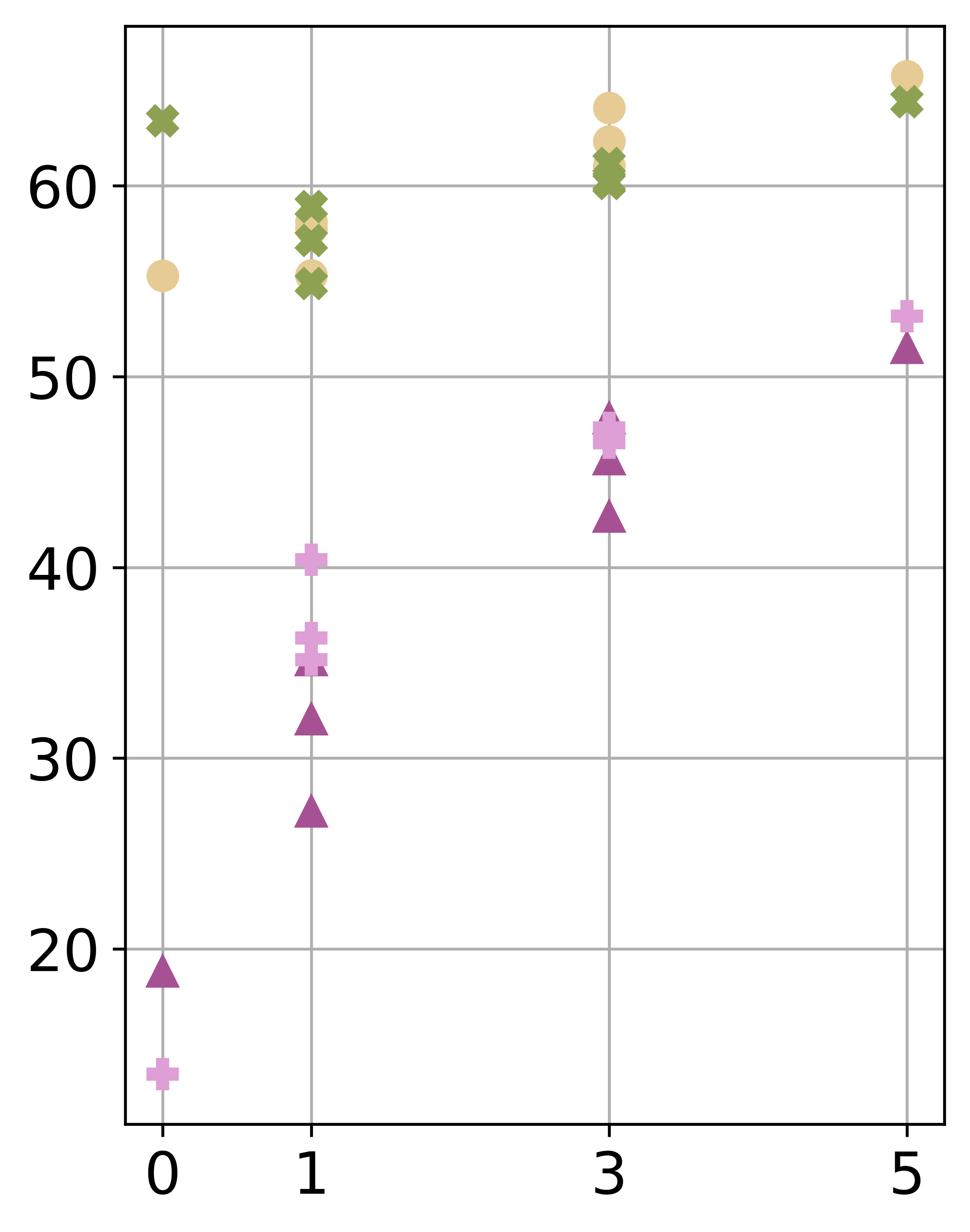}
        \caption{QA template}
        \label{fig:correlation-fewshot-ave-qa}
    \end{subfigure}
    \hfill
    \begin{subfigure}[b]{0.47\columnwidth}
        \centering
        \includegraphics[width=\columnwidth]{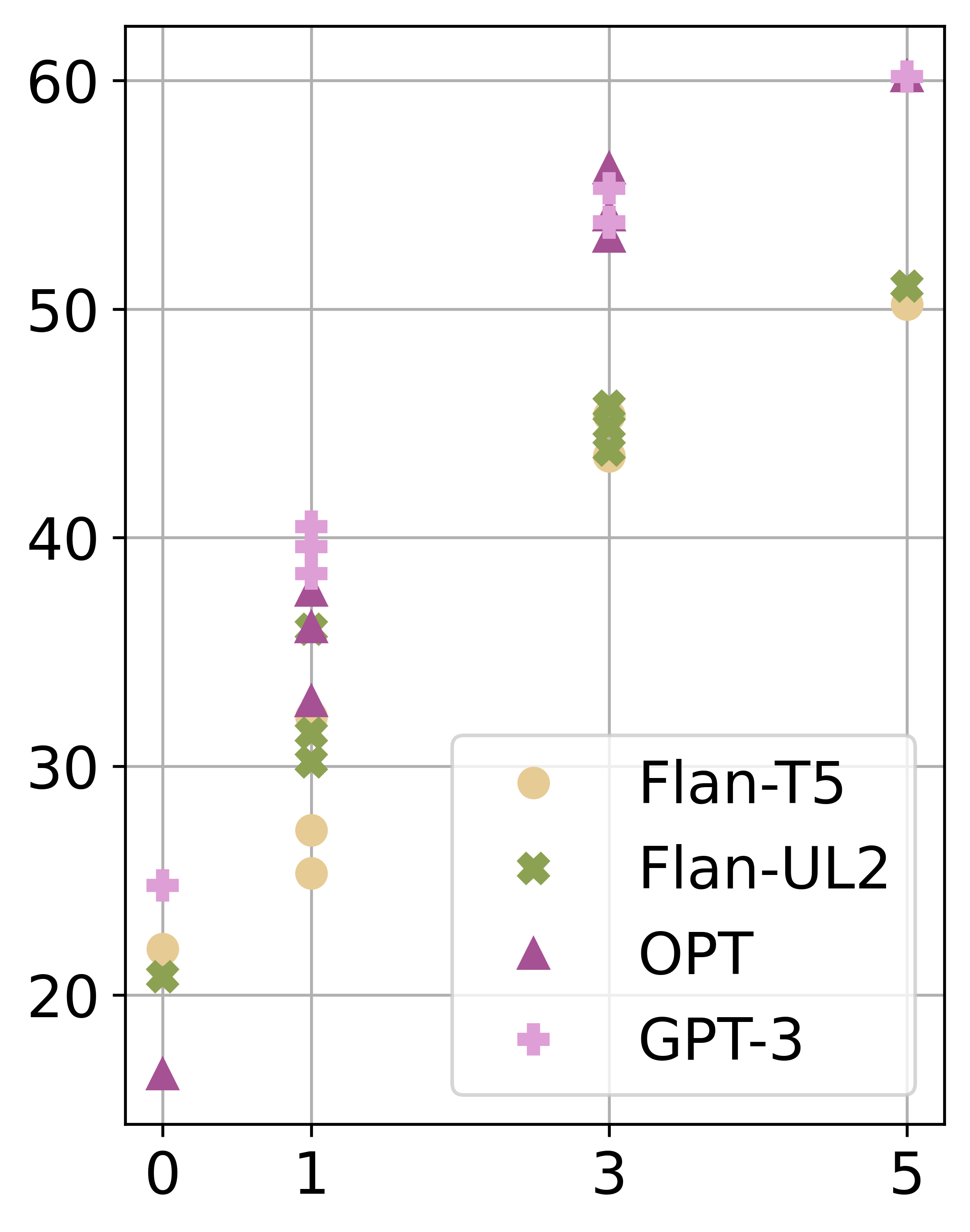}
        \caption{LC template}
        \label{fig:correlation-fewshot-ave-lc}
    \end{subfigure}
    \caption{Spearman's rank correlation averaged over the five relation types with different number of the prototypical examples. For 1-shot and 3-shot examples, we report each correlation of the three individual runs.}
    \label{fig:correlation-fewshot-ave}
\end{figure}

\autoref{fig:correlation-fewshot-ave-qa} shows the results for the QA template. We can see that all models improve when more prototypical examples are provided, with the zero-shot performance of Flan-UL2 being an outlier. Remarkably, Flan-UL2 achieves 62.5\% accuracy in the zero-shot setting, which is competitive with the 5-shot results in \autoref{tab:baselines}. Flan-T5\textsubscript{XXL} also achieves a zero-shot result of 54.5\%, which is better than most of the models in the main (5-shot) experiments. In the zero-shot setting, OPT\textsubscript{13B} performs better than GPT-3\textsubscript{davinci}, but GPT-3\textsubscript{davinci} quickly improves as more examples are provided, clearly outperforming OPT\textsubscript{13B} in the 5-shot setting. \autoref{fig:correlation-fewshot-ave-lc} shows the results for the LC template. We again see that providing more examples benefits all models.
Unlike for the QA template, however, Flan-T5\textsubscript{XXL} performs poorly in the zero-shot setting. Moreover, OPT\textsubscript{13B} now sees the largest improvement between the zero-shot and 5-shot settings.

\subsection{Qualitative Analysis}\label{secQualitative}
To better understand the predictions of the models, we analyse the most flagrant mistakes. Specifically, we focus on those entity pairs whose predicted rank is in the top 30\%, while being in the bottom 30\% of the gold ranking, and vice versa. \autoref{tab:qualitative-top} and \autoref{tab:qualitative-bottom} show the entity pairs from the test set for which this was the case. For this analysis, we look at the models with their optimal templates: i.e., Flan-T5 and Flan-UL2 with the QA template, and the other models with the LC template.

\begin{table}[!t]
\centering
\footnotesize
\begin{tabular}{@{}l@{\hspace{5pt}}l@{\hspace{5pt}}p{180pt}@{}}
\toprule
&  & \textbf{Incorrectly predicted to be in the top 30\%}  \\ \midrule
\multirow{8}{*}{\rotatebox{90}{Flan-T5\textsubscript{XXL}}} &     \multirow{2}{*}{Ally} &                                        Armenia : Azerbaijan, Liam Gallagher : Noel Gallagher, Russia : Georgia \\ \cmidrule{3-3}
 &      \multirow{3}{*}{Inf} & Harry Potter : Wizard of Oz, heavy metal : punk music, Luke Bryan : Hank Williams, James Brown : Michael Jackson \\ \cmidrule{3-3}
 &      \multirow{2}{*}{Sim} &                                                          sphinx : sphynx, New York : York, cannoli : canneloni \\ \midrule
\multirow{9}{*}{\rotatebox{90}{Flan-UL2}} &    Rival &                                                                              Serena Williams : Andy Murray \\\cmidrule{3-3}
 &     \multirow{2}{*}{Ally} &                                                            Liam Gallagher : Noel Gallagher, Google : Samsung \\\cmidrule{3-3}
 &     \multirow{2}{*}{Inf} &                           Harry Potter : Wizard of Oz, heavy metal : punk music, James Brown : Michael Jackson \\\cmidrule{3-3}
 &     Know &                                                                                             Belgium : wine \\\cmidrule{3-3}
 &      Sim &                                                                         sphinx : sphynx, cannoli : canneloni \\ \midrule
\multirow{8}{*}{\rotatebox{90}{OPT\textsubscript{13B}}} &    Rival &                                                                              Serena Williams : Andy Murray \\\cmidrule{3-3}
 &     \multirow{2}{*}{Ally} &                            Joseph Stalin : Josip Broz Tito, Armenia : Azerbaijan, Sophia Loren : Marlon Brando \\\cmidrule{3-3}
 &      \multirow{2}{*}{Inf} &                       Joe Biden : Donald Trump, Harry Potter : Wizard of Oz, Singaporean food : Malaysian food \\\cmidrule{3-3}
 &     Know &                                                                      Coca-Cola : Pepsi, Steve Jobs : AirPods \\ \midrule
\multirow{8}{*}{\rotatebox{90}{GPT-3\textsubscript{davinci}}} &    Rival &                                                                              Serena Williams : Andy Murray \\\cmidrule{3-3}
 &     \multirow{2}{*}{Ally} &                         Joseph Stalin : Josip Broz Tito, Armenia : Azerbaijan, Liam Gallagher : Noel Gallagher \\\cmidrule{3-3}
 &      Inf &                                                                                Harry Potter : Wizard of Oz \\\cmidrule{3-3}
 &     Know &                                                                                          Coca-Cola : Pepsi \\\cmidrule{3-3}
 &      Sim &                                                                       Nicolae Ceaușescu : Javier Hernández \\
\bottomrule
\end{tabular}
\caption{Test examples of incorrect predictions made by the three best models in the top 30\%.}
\label{tab:qualitative-top}
\end{table}

\begin{table}[!t]
\centering
\footnotesize
\begin{tabular}{@{}l@{\hspace{5pt}}l@{\hspace{5pt}}p{180pt}@{}}
\toprule
 &  & \textbf{Incorrectly predicted to be in the bottom 30\%}  \\ \midrule
\multirow{9}{*}{\rotatebox{90}{Flan-T5\textsubscript{XXL}}} &    Rival &                                                                                    Isaac Newton : Gottfried Leibniz \\\cmidrule{3-3}
 &     \multirow{2}{*}{Ally} &                                                   China : North Korea, Ron Weasley : Neville Longbottom, Windows : Xbox \\\cmidrule{3-3}
 &      \multirow{3}{*}{Inf} & Prince Harry : Monarchy, trending music : TikTok, Coca-Cola : Pepsi, Apple Music : Spotify, Pepsi : Coca-Cola, Hoover : Dyson \\\cmidrule{3-3}
 &     Know &                                                                         Corsica : Napoleon Bonaparte, France : cheese \\\cmidrule{3-3}
 &      Sim &                                                                                     Suits : Law\&Order, Shark : Bush \\ \midrule
\multirow{10}{*}{\rotatebox{90}{Flan-UL2}} &     \multirow{3}{*}{Ally} &                         Tata Motors : Jaguar, China : North Korea, HSBC : BlackRock, Coca-Cola : McDonald's, Huawei : China \\\cmidrule{3-3}
 &      \multirow{2}{*}{Inf} &                        Prince Harry : Monarchy, trending music : TikTok, Wales : Westminster, Theresa May : David Cameron \\\cmidrule{3-3}
 &     \multirow{2}{*}{Know} &                                            Europe : The Final Countdown, Corsica : Napoleon Bonaparte, OpenAI : ChatGPT \\\cmidrule{3-3}
 &      \multirow{2}{*}{Sim} &                                                             Minnesota : Wisconsin, Shark : Bush, Glastonbury : Roskilde \\ \midrule
\multirow{9}{*}{\rotatebox{90}{OPT\textsubscript{13B}}}  &     \multirow{3}{*}{Ally} &                            FTX : Alameda Research, Red Bull : GoPro, HSBC : BlackRock, Microsoft : LinkedIn, Windows : Xbox \\\cmidrule{3-3}
 &      \multirow{2}{*}{Inf} &                                                   Prince Harry : Monarchy, trending music : TikTok, Wales : Westminster \\\cmidrule{3-3}
 &     Know &                                                                                           OpenAI : ChatGPT, UK : rain \\\cmidrule{3-3}
 &      \multirow{2}{*}{Sim} &                                             pill : tablet, Great Britian : British Empire, fusilli : rotini, Shark : Bush \\ \midrule
\multirow{9}{*}{\rotatebox{90}{GPT-3\textsubscript{davinci}}} &    Rival &                                                                                               Netflix : Disney Plus \\\cmidrule{3-3}
 &     \multirow{2}{*}{Ally} &                                     FTX : Alameda Research, Rishi Sunak : Joe Biden, Microsoft : LinkedIn, Windows : Xbox \\\cmidrule{3-3}
 &      \multirow{2}{*}{Inf} &                                          Prince Harry : Monarchy, trending music : TikTok, Stephen King : Arthur Machen \\\cmidrule{3-3}
 &     Know &                                                                                                    OpenAI:ChatGPT \\\cmidrule{3-3}
 &      \multirow{2}{*}{Sim} &                                                          Homebase : IKEA, fusilli : rotini, Shark : Bush, Primark : Shein \\
\bottomrule
\end{tabular}
\caption{Test examples of incorrect predictions made by the three best models in the bottom 30\%.}
\label{tab:qualitative-bottom}
\end{table}

When looking at the instances that mistakenly end up in the top 30\%, we see entities which are closely related (e.g.\ ``Coca-Cola : Pepsi'') while not actually satisfying the intended relation. We can see several cases where entities with similar names are mistakenly predicted to be similar (e.g.\ sphinx : sphynx, New York : York, cannoli : canneloni). Several models also mistakenly predict ``Serena Williams : Andy Murray'' as an instance of the rival-of relation, presumably because the model has learned that players from the same sport are often rivals.
When looking at the examples from the bottom 30\%, we can see entities which only recently became prominent (e.g.\ FTX and Alameda Research), highlighting the limitation of using language models that have not been trained on the most recent data. The ``Corsica : Napoleon Bonaparte'', ``Prince Harry : Monarchy'' and ``trending music : TikTok'' examples illustrate how the models can struggle with cases involving entities of different semantic types. 

\section{Conclusions}
In this paper, we have proposed the task of modelling graded relations between named entities, with a new dataset. The task consists in ranking entity pairs according to how much they satisfy a given graded relation, where models only have access to the description of the relation and five prototypical instances per relation. To assess the difficulty of the task, we analysed a large number of baselines, including public LLMs of up to 30B parameters, state-of-the-art relation embedding models, and closed LLMs such as GPT-4. We found significant performance differences between the largest LMs and their smaller siblings, which highlights the progress achieved in NLP in the last few years by scaling up LMs. However, even the largest models trail human performance by around 15 percentage points.

\section*{Limitations}
Our dataset is aimed at testing the ability of LMs to understand graded relations between named entities. In particular, the size of the dataset makes it unsuitable for training models (beyond the few-shot setting). Furthermore, our dataset is limited to five relation types. We believe these relations to be among the most prominent graded relations between named entities. Nonetheless, there are clearly various other relations that could be considered, especially in domain-specific settings. While the annotation process involved comprehensive quality control mechanisms, the dataset may have inherited some of the biases of the annotators. The annotators were diverse in terms of gender, nationality and cultural background, but all came from the the same academic setting. Since the annotation is inherently subjective, this may be reflected in the final dataset. Finally, the task may have a temporal component in which some relationships may change over time. Our annotations represents the views of the annotators at a particular moment in time. In future, the dataset could be extended, to provide different temporal snapshots, which would allow an evaluation of ability of LMs to model temporal context.

\section*{Ethics Statement}
Our data has been created and labelled by human annotators. As such, we have ensured that proper training was provided, and that annotators were paid fairly through our institutional student job provider. We also acknowledge the potential biases of our dataset, and the potentially sensitive nature of examples related to political or religious content. To mitigate this issue, we have relied on a diverse set of annotators, and we have provided guidelines about avoiding sensitive content.

\section*{Acknowledgements}

Jose Camacho-Collados is supported by a UKRI Future Leaders Fellowship. Steven Schockaert was supported by EPSRC grant EP/V025961/1. We thank all the annotators for their help in the construction of the dataset.

\bibliography{anthology,custom}

\appendix

\section{Annotator Agreement for Each Relation}
\label{app:iaa}

\begin{table}[!t]
\centering
\footnotesize
\begin{tabular}{c@{\hspace{7pt}}c@{\hspace{7pt}}c@{\hspace{7pt}}c@{\hspace{7pt}}c@{\hspace{7pt}}c@{\hspace{7pt}}c@{\hspace{7pt}}c@{\hspace{7pt}}c}
\toprule
{} &    A &    B &    C &    D &    E &    F &    G &  Others \\
\midrule
A   &  100 &   53 &   77 &   63 &   64 &   68 &   67 &      80 \\
B   &   53 &  100 &   52 &   43 &   47 &   46 &   48 &      56 \\
C   &   77 &   52 &  100 &   63 &   58 &   67 &   68 &      79 \\
D   &   63 &   43 &   63 &  100 &   48 &   54 &   59 &      66 \\
E   &   64 &   47 &   58 &   48 &  100 &   57 &   59 &      65 \\
F   &   68 &   46 &   67 &   54 &   57 &  100 &   62 &      70 \\
G   &   67 &   48 &   68 &   59 &   59 &   62 &  100 &      73 \\ \midrule
AVG &   70 &   55 &   69 &   61 &   62 &   65 &   66 &      70 \\
\bottomrule
\end{tabular}
\caption{Spearman correlation (\%) between each pair of annotators (A,...,G), and between each annotator and the average score provided by the other six averaged over all the five relation types \textbf{before the 3rd and final quality enhancement annotation round.}}
\label{tab:annotator-agreement-before}
\end{table}

\begin{table}[!t]
\centering
\footnotesize
\begin{tabular}{@{}c@{\hspace{7pt}}c@{\hspace{7pt}}c@{\hspace{7pt}}c@{\hspace{7pt}}c@{\hspace{7pt}}c@{\hspace{7pt}}c@{\hspace{7pt}}c@{\hspace{7pt}}c@{}}
\toprule
{} &    A &    B &    C &    D &    E &    F &    G &  Others \\
\midrule
A   &  100 &   55 &   79 &   69 &   74 &   78 &   79 &      86 \\
B   &   55 &  100 &   46 &   35 &   58 &   57 &   50 &      54 \\
C   &   79 &   46 &  100 &   75 &   67 &   73 &   75 &      80 \\
D   &   69 &   35 &   75 &  100 &   52 &   66 &   68 &      74 \\
E   &   74 &   58 &   67 &   52 &  100 &   69 &   67 &      74 \\
F   &   78 &   57 &   73 &   66 &   69 &  100 &   65 &      79 \\
G   &   79 &   50 &   75 &   68 &   67 &   65 &  100 &      79 \\ \midrule
AVG &   76 &   57 &   74 &   66 &   70 &   73 &   72 &      75 \\
\bottomrule
\end{tabular}
\caption{Spearman correlation (\%) on the \emph{competitor/rival of} relation between each pair of annotators (A,...,G), and between each annotator and the average score provided by the other six \textbf{after the 3rd and final quality enhancement annotation round.}}
\label{tab:annotator-agreement-rival-after}
\end{table}

\begin{table}[!t]
\centering
\footnotesize
\begin{tabular}{@{}c@{\hspace{7pt}}c@{\hspace{7pt}}c@{\hspace{7pt}}c@{\hspace{7pt}}c@{\hspace{7pt}}c@{\hspace{7pt}}c@{\hspace{7pt}}c@{\hspace{7pt}}c@{}}
\toprule
{} &    A &    B &    C &    D &    E &    F &    G &  Others \\
\midrule
A   &  100 &   73 &   85 &   69 &   74 &   78 &   73 &      87 \\
B   &   73 &  100 &   74 &   52 &   64 &   72 &   65 &      75 \\
C   &   85 &   74 &  100 &   68 &   72 &   77 &   74 &      87 \\
D   &   69 &   52 &   68 &  100 &   63 &   59 &   65 &      69 \\
E   &   74 &   64 &   72 &   63 &  100 &   67 &   70 &      76 \\
F   &   78 &   72 &   77 &   59 &   67 &  100 &   75 &      80 \\
G   &   73 &   65 &   74 &   65 &   70 &   75 &  100 &      78 \\ \midrule
AVG &   79 &   71 &   78 &   68 &   73 &   76 &   75 &      79 \\
\bottomrule
\end{tabular}
\caption{Spearman correlation (\%) on the \emph{friend/ally of} relation between each pair of annotators (A,...,G), and between each annotator and the average score provided by the other six \textbf{after the 3rd and final quality enhancement annotation round.}}
\label{tab:annotator-agreement-ally-after}
\end{table}

\begin{table}[!t]
\centering
\footnotesize
\begin{tabular}{@{}c@{\hspace{7pt}}c@{\hspace{7pt}}c@{\hspace{7pt}}c@{\hspace{7pt}}c@{\hspace{7pt}}c@{\hspace{7pt}}c@{\hspace{7pt}}c@{\hspace{7pt}}c@{}}
\toprule
{} &    A &    B &    C &    D &    E &    F &    G &  Others \\
\midrule
A   &  100 &   50 &   76 &   68 &   69 &   59 &   71 &      76 \\
B   &   50 &  100 &   55 &   63 &   49 &   32 &   54 &      55 \\
C   &   76 &   55 &  100 &   74 &   70 &   69 &   76 &      84 \\
D   &   68 &   63 &   74 &  100 &   65 &   52 &   70 &      76 \\
E   &   69 &   49 &   70 &   65 &  100 &   65 &   71 &      71 \\
F   &   59 &   32 &   69 &   52 &   65 &  100 &   62 &      61 \\
G   &   71 &   54 &   76 &   70 &   71 &   62 &  100 &      78 \\ \midrule
AVG &   70 &   58 &   74 &   70 &   70 &   63 &   72 &      71 \\
\bottomrule
\end{tabular}
\caption{Spearman correlation (\%) on the \emph{influenced by} relation between each pair of annotators (A,...,G), and between each annotator and the average score provided by the other six \textbf{after the 3rd and final quality enhancement annotation round.}}
\label{tab:annotator-agreement-inf-after}
\end{table}

\begin{table}[!t]
\centering
\footnotesize
\begin{tabular}{@{}c@{\hspace{7pt}}c@{\hspace{7pt}}c@{\hspace{7pt}}c@{\hspace{7pt}}c@{\hspace{7pt}}c@{\hspace{7pt}}c@{\hspace{7pt}}c@{\hspace{7pt}}c@{}}
\toprule
{} &    A &    B &    C &    D &    E &    F &    G &  Others \\
\midrule
A   &  100 &   74 &   84 &   78 &   80 &   80 &   77 &      88 \\
B   &   74 &  100 &   71 &   70 &   73 &   65 &   70 &      76 \\
C   &   84 &   71 &  100 &   77 &   77 &   75 &   80 &      88 \\
D   &   78 &   70 &   77 &  100 &   76 &   82 &   75 &      83 \\
E   &   80 &   73 &   77 &   76 &  100 &   71 &   76 &      81 \\
F   &   80 &   65 &   75 &   82 &   71 &  100 &   71 &      80 \\
G   &   77 &   70 &   80 &   75 &   76 &   71 &  100 &      82 \\ \midrule
AVG &   82 &   75 &   81 &   80 &   79 &   78 &   78 &      83 \\
\bottomrule
\end{tabular}
\caption{Spearman correlation (\%) on the \emph{known for} relation between each pair of annotators (A,...,G), and between each annotator and the average score provided by the other six \textbf{after the 3rd and final quality enhancement annotation round.}}
\label{tab:annotator-agreement-know-after}
\end{table}

\begin{table}[!t]
\centering
\footnotesize
\begin{tabular}{@{}c@{\hspace{7pt}}c@{\hspace{7pt}}c@{\hspace{7pt}}c@{\hspace{7pt}}c@{\hspace{7pt}}c@{\hspace{7pt}}c@{\hspace{7pt}}c@{\hspace{7pt}}c@{}}
\toprule
{} &    A &    B &    C &    D &    E &    F &    G &  Others \\
\midrule
A   &  100 &   58 &   82 &   74 &   79 &   78 &   73 &      82 \\
B   &   58 &  100 &   61 &   64 &   64 &   59 &   61 &      68 \\
C   &   82 &   61 &  100 &   74 &   75 &   74 &   70 &      79 \\
D   &   74 &   64 &   74 &  100 &   77 &   77 &   73 &      83 \\
E   &   79 &   64 &   75 &   77 &  100 &   75 &   78 &      84 \\
F   &   78 &   59 &   74 &   77 &   75 &  100 &   74 &      79 \\
G   &   73 &   61 &   70 &   73 &   78 &   74 &  100 &      78 \\ \midrule
AVG &   78 &   67 &   76 &   77 &   78 &   77 &   75 &      79 \\
\bottomrule
\end{tabular}
\caption{Spearman correlation (\%) on the \emph{similar to} relation between each pair of annotators (A,...,G), and between each annotator and the average score provided by the other six \textbf{after the 3rd and final quality enhancement annotation round.}}
\label{tab:annotator-agreement-sim-after}
\end{table}

\autoref{tab:annotator-agreement-before} show the Spearman correlation for each relation type as well as the average over all the relation types before the 3rd and final quality enhancement annotation round.
\autoref{tab:annotator-agreement-rival-after},
\autoref{tab:annotator-agreement-ally-after}, 
\autoref{tab:annotator-agreement-inf-after},
\autoref{tab:annotator-agreement-know-after}, and
\autoref{tab:annotator-agreement-sim-after} show the Spearman correlation for each relation type after the 3rd and final quality enhancement annotation round.

\section{Models on HuggingFace}
\label{app:hf-name}

\begin{table}[t]
\centering
\footnotesize
\begin{tabular}{@{}ll@{}}
\toprule
Model &        Name on HuggingFace \\
\midrule
RelBERT\textsubscript{BASE}     &               \texttt{relbert/relbert-roberta-base} \\
RelBERT\textsubscript{LARGE}    &              \texttt{relbert/relbert-roberta-large} \\
\midrule
OPT\textsubscript{125M}     &      \texttt{facebook/opt-125m} \\
OPT\textsubscript{350M}     &      \texttt{facebook/opt-350m} \\
OPT\textsubscript{1.3B}     &      \texttt{facebook/opt-1.3b} \\
OPT\textsubscript{2.7B}     &      \texttt{facebook/opt-2.7b} \\
OPT\textsubscript{6.7B}     &      \texttt{facebook/opt-6.7b} \\
OPT\textsubscript{13B}     &      \texttt{facebook/opt-13b} \\
OPT\textsubscript{30B}     &      \texttt{facebook/opt-30b} \\
OPT\textsubscript{66B}     &      \texttt{facebook/opt-66b} \\
\midrule
OPT-IML\textsubscript{1.3B}     &      \texttt{facebook/opt-iml-1.3b} \\
OPT-IML\textsubscript{30B}      &   \texttt{facebook/opt-iml-30b} \\
OPT-IML\textsubscript{MAX-1.3B} &  \texttt{facebook/opt-iml-max-1.3b} \\
OPT-IML\textsubscript{MAX-30B}      &   \texttt{facebook/opt-iml-max-30b} \\
\midrule
T5\textsubscript{SMALL}    &       \texttt{t5-small} \\
T5\textsubscript{BASE}     &        \texttt{t5-base} \\
T5\textsubscript{LARGE}    &       \texttt{t5-large} \\
T5\textsubscript{XL}       &          \texttt{t5-3b} \\
T5\textsubscript{XXL}      &         \texttt{t5-11b} \\
\midrule
Flan-T5\textsubscript{SMALL}    &       \texttt{google/flan-t5-small} \\
Flan-T5\textsubscript{BASE}     &        \texttt{google/flan-t5-base} \\
Flan-T5\textsubscript{LARGE}    &       \texttt{google/flan-t5-large} \\
Flan-T5\textsubscript{XL}       &          \texttt{google/flan-t5-xl} \\
Flan-T5\textsubscript{XXL}      &         \texttt{google/flan-t5-xxl} \\
Flan-UL2\textsubscript{20B}     &            \texttt{google/flan-ul2} \\
\bottomrule
\end{tabular}
\caption{The language models used in the paper and their corresponding alias on HuggingFace model hub.}
\label{tab:hf-name}
\end{table}

\autoref{tab:hf-name} shows the model alias on the HuggingFace of the LMs we used in our experiment.

\section{Additional Results}
\label{app:additional-result}

\begin{figure}[!t]
    \centering
    \begin{subfigure}[b]{0.47\columnwidth}
        \centering
        \includegraphics[width=\columnwidth]{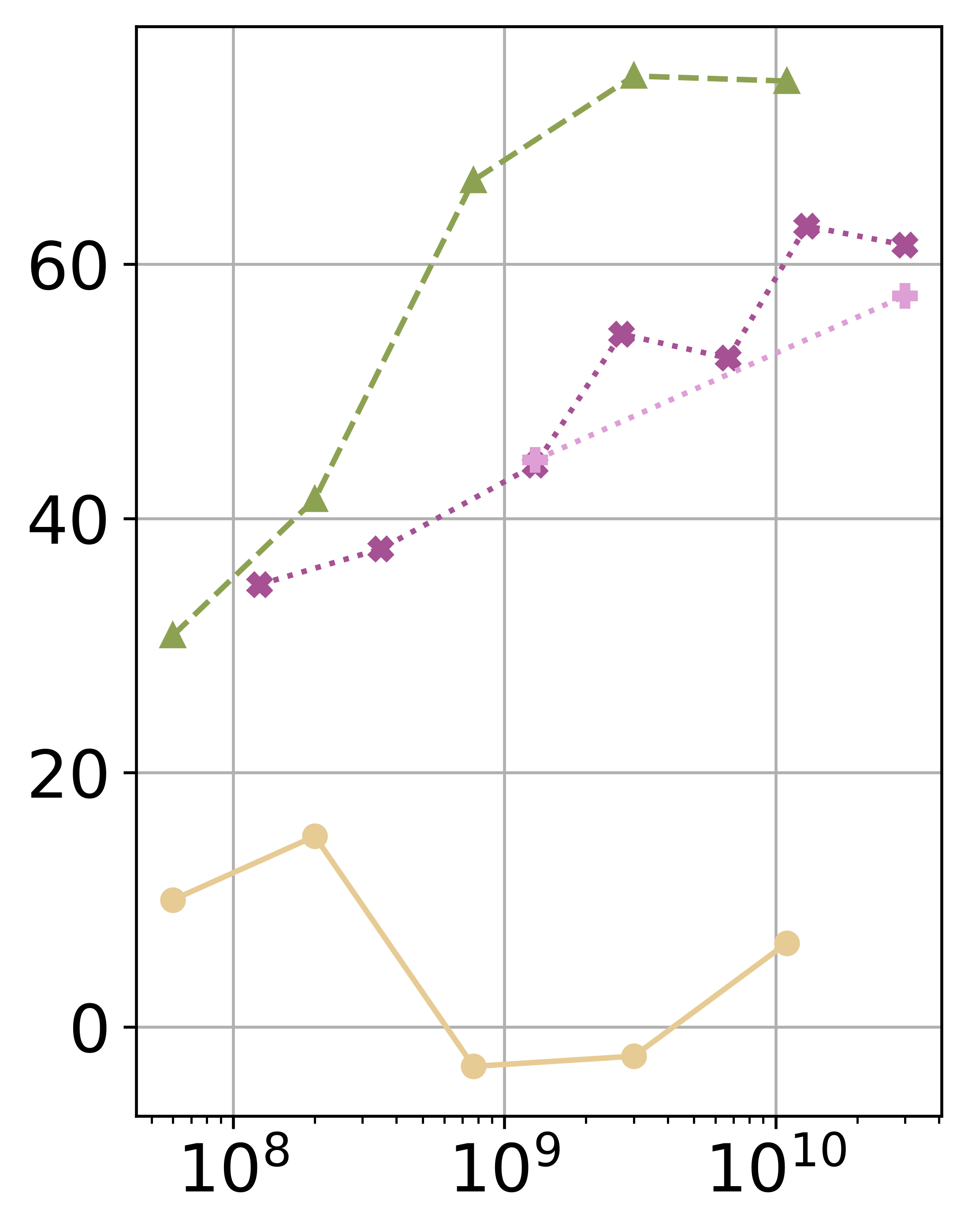}
        \caption{QA template}
    \end{subfigure}     
    \begin{subfigure}[b]{0.47\columnwidth}
        \centering
        \includegraphics[width=\columnwidth]{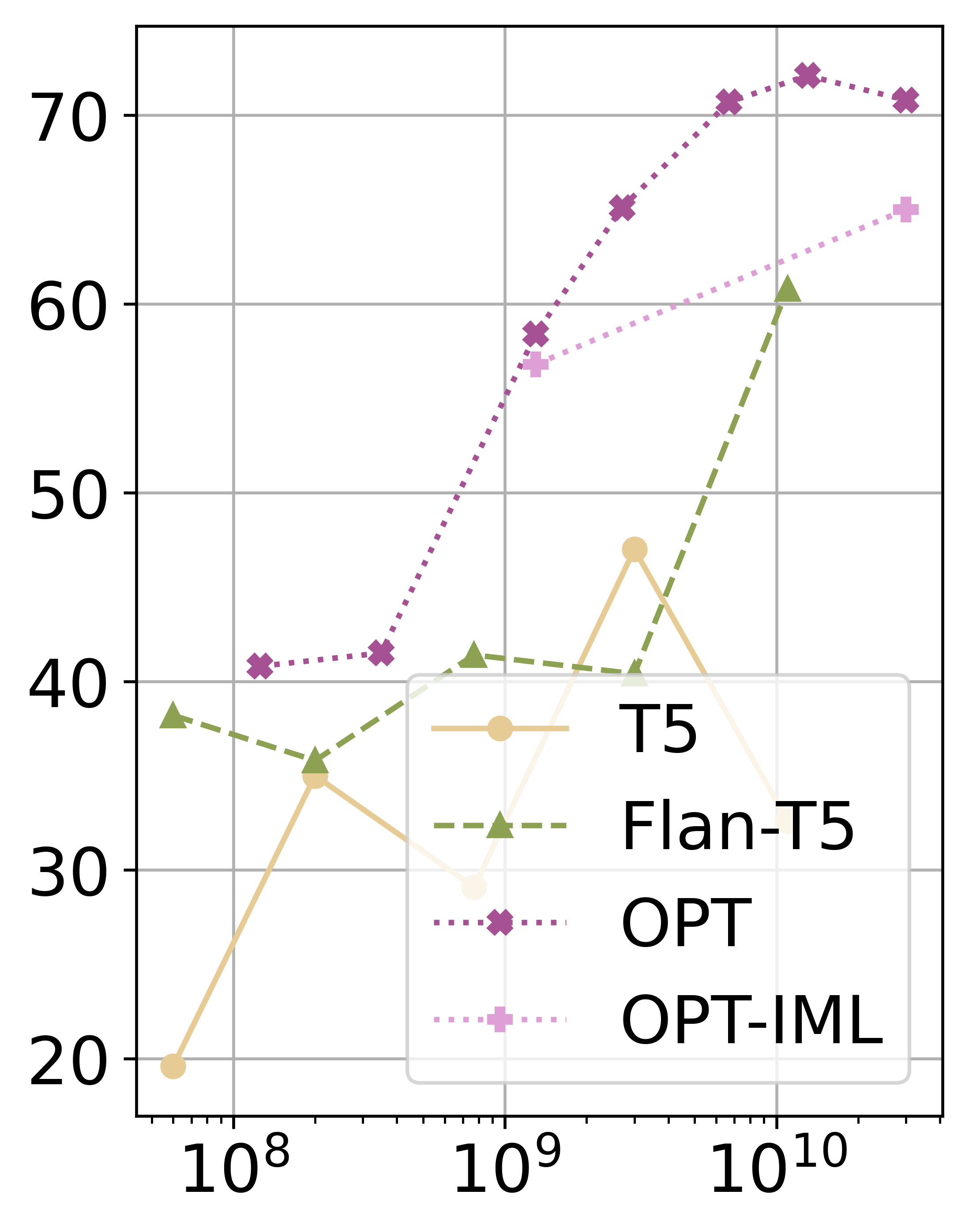}
        \caption{LC template}
    \end{subfigure}
    \caption{Spearman's rank correlation for the \emph{competitor/rival of} relation type along with the model size.}
    \label{fig:correlation-with-model-size-rival}
\end{figure}

\begin{figure}[!t]
    \centering
    \begin{subfigure}[b]{0.47\columnwidth}
        \centering
        \includegraphics[width=\columnwidth]{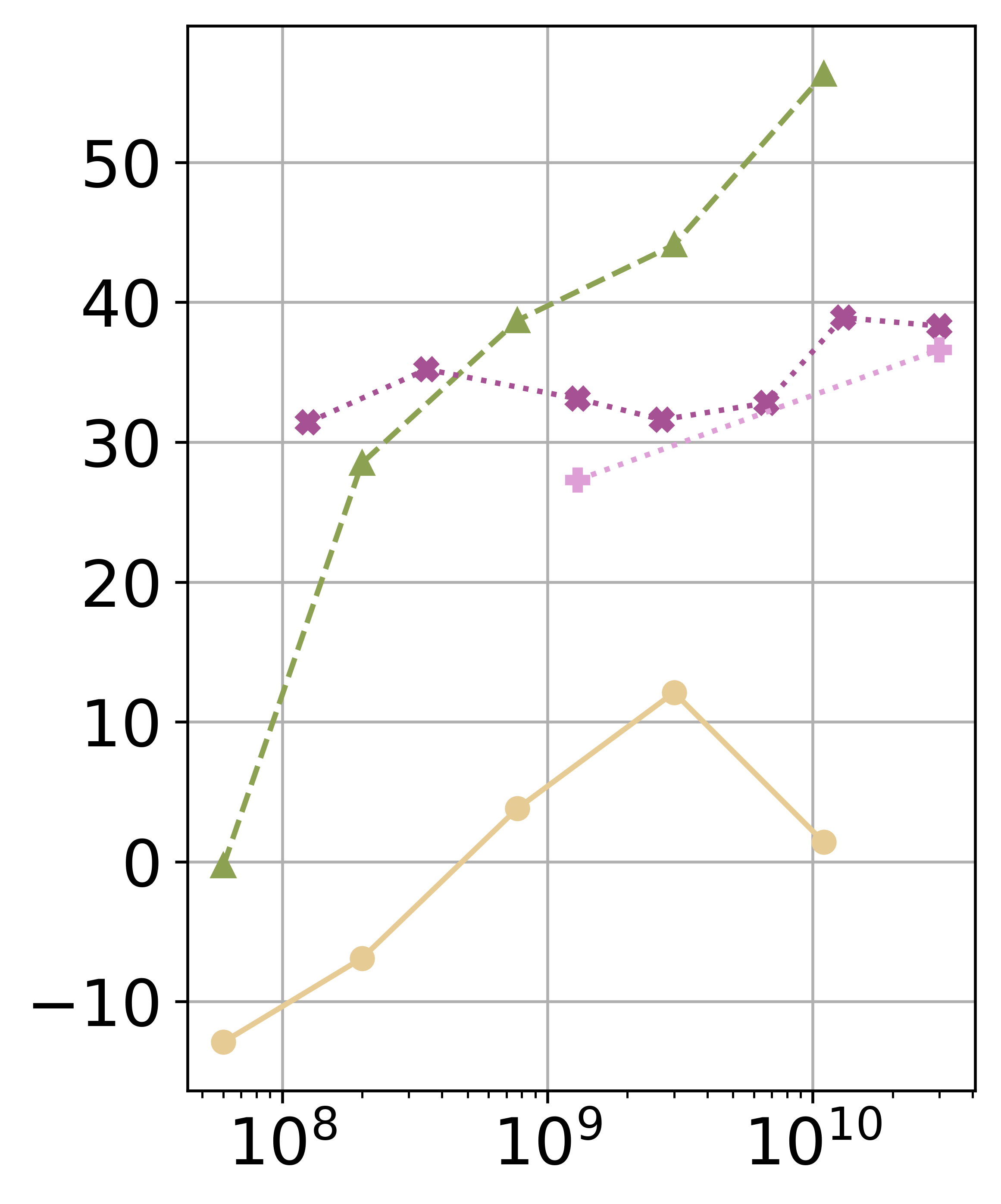}
        \caption{QA template}
    \end{subfigure}     
    \begin{subfigure}[b]{0.47\columnwidth}
        \centering
        \includegraphics[width=\columnwidth]{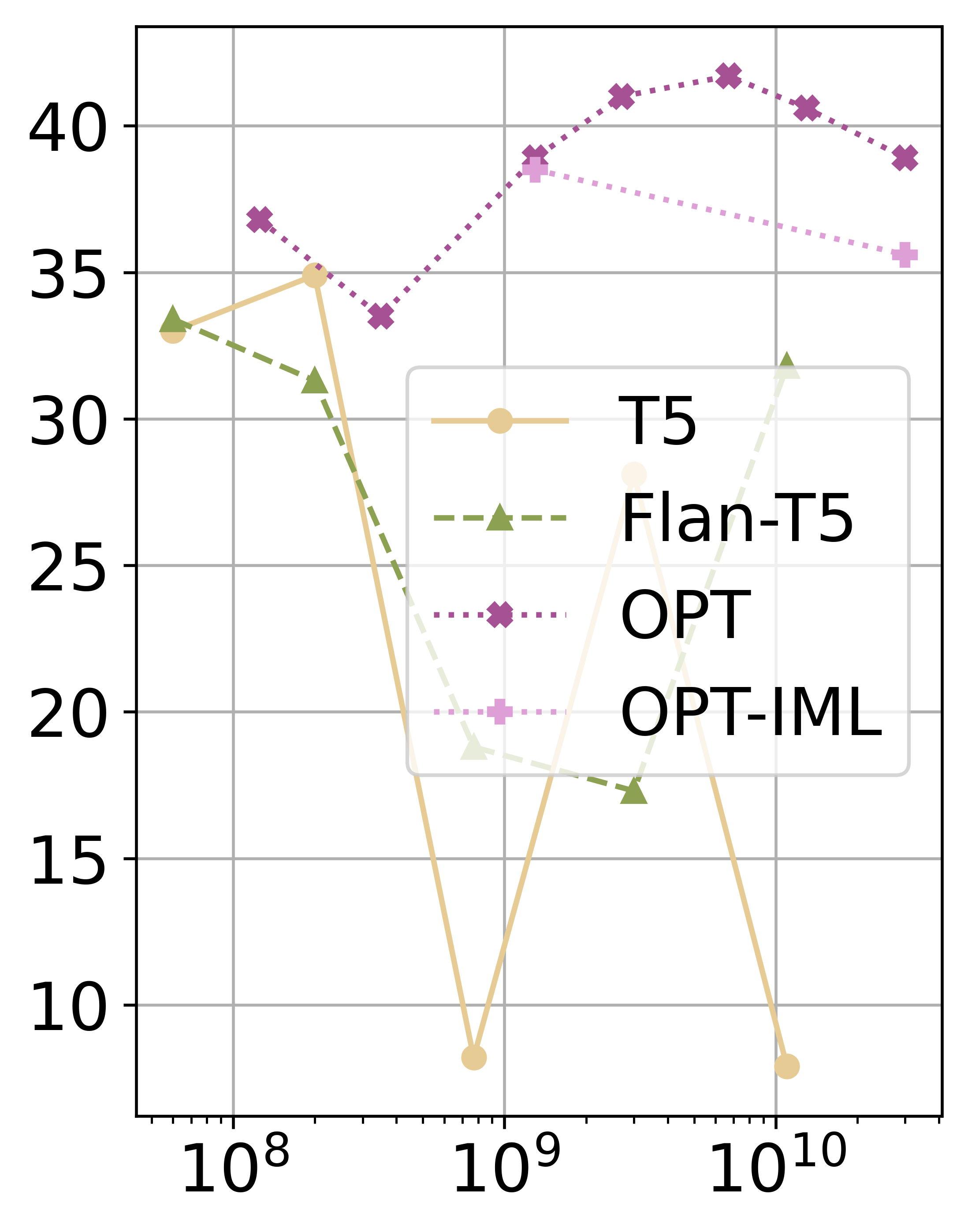}
        \caption{LC template}
    \end{subfigure}
    \caption{Spearman's rank correlation for the \emph{friend/ally of} relation type along with the model size.}
    \label{fig:correlation-with-model-size-ally}
\end{figure}

\begin{figure}[!t]
    \centering
    \begin{subfigure}[b]{0.47\columnwidth}
        \centering
        \includegraphics[width=\columnwidth]{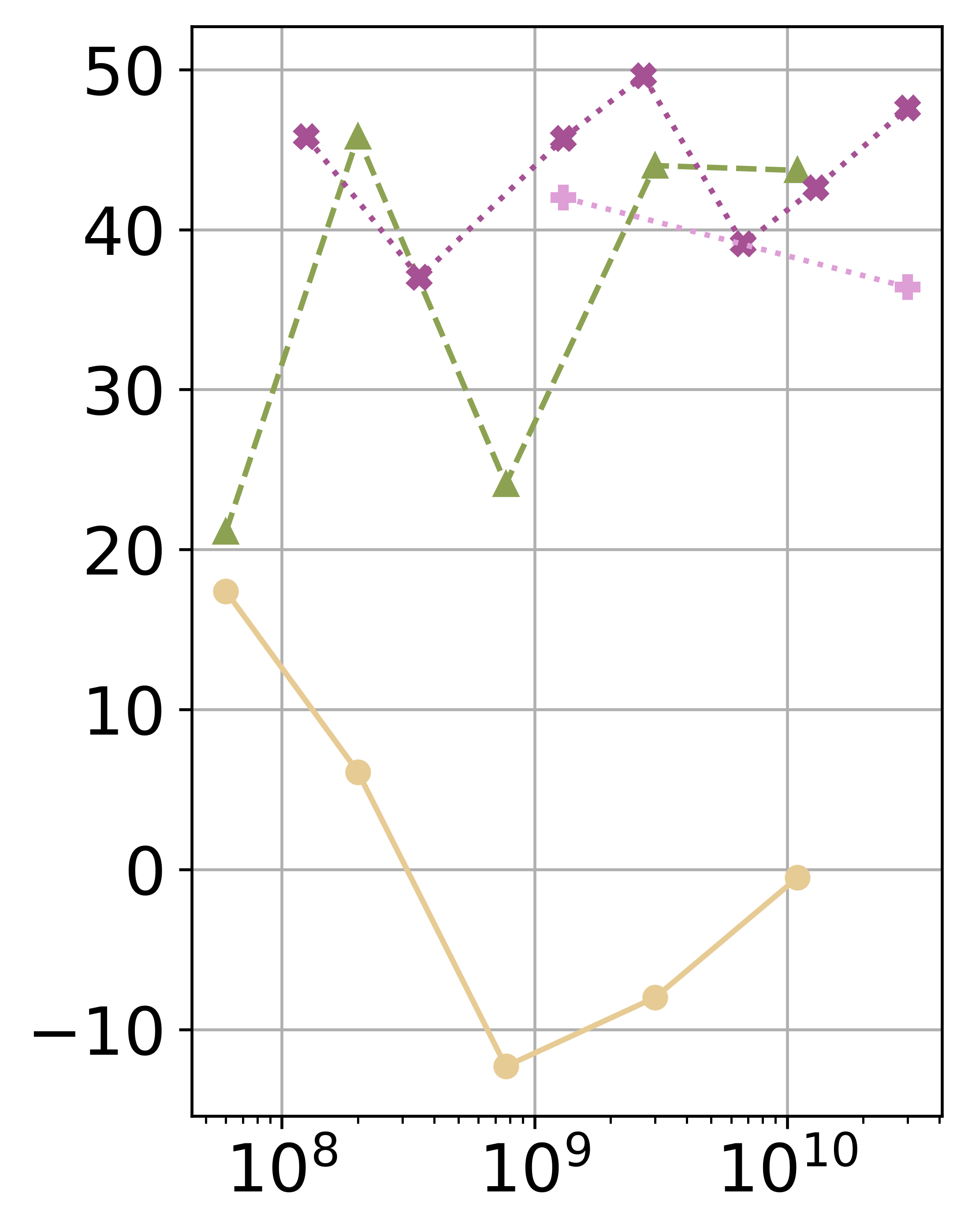}
        \caption{QA template}
    \end{subfigure}     
    \begin{subfigure}[b]{0.47\columnwidth}
        \centering
        \includegraphics[width=\columnwidth]{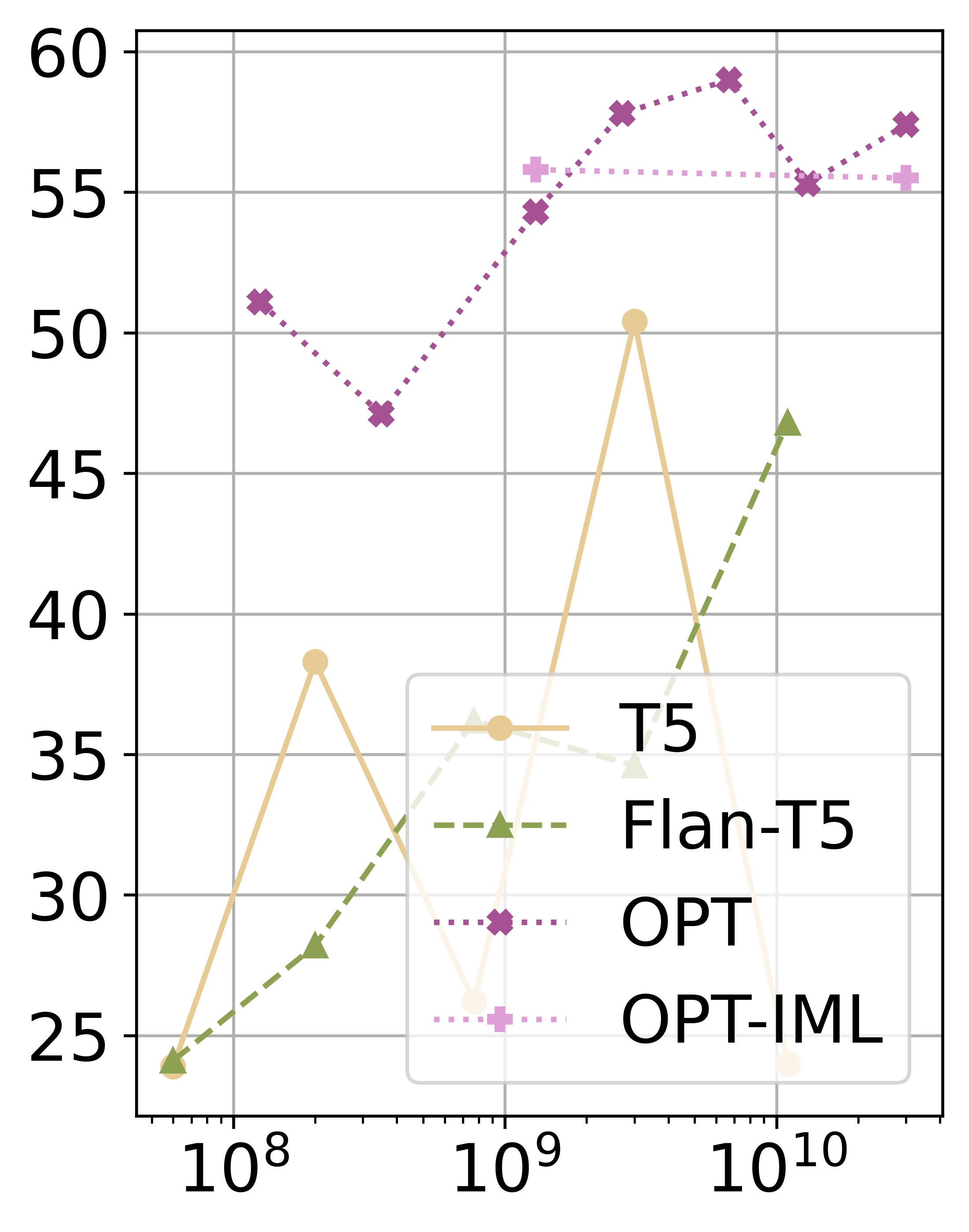}
        \caption{LC template}
    \end{subfigure}
    \caption{Spearman's rank correlation for the \emph{influenced by} relation type along with the model size.}
    \label{fig:correlation-with-model-size-inf}
\end{figure}

\begin{figure}[!t]
    \centering
    \begin{subfigure}[b]{0.47\columnwidth}
        \centering
        \includegraphics[width=\columnwidth]{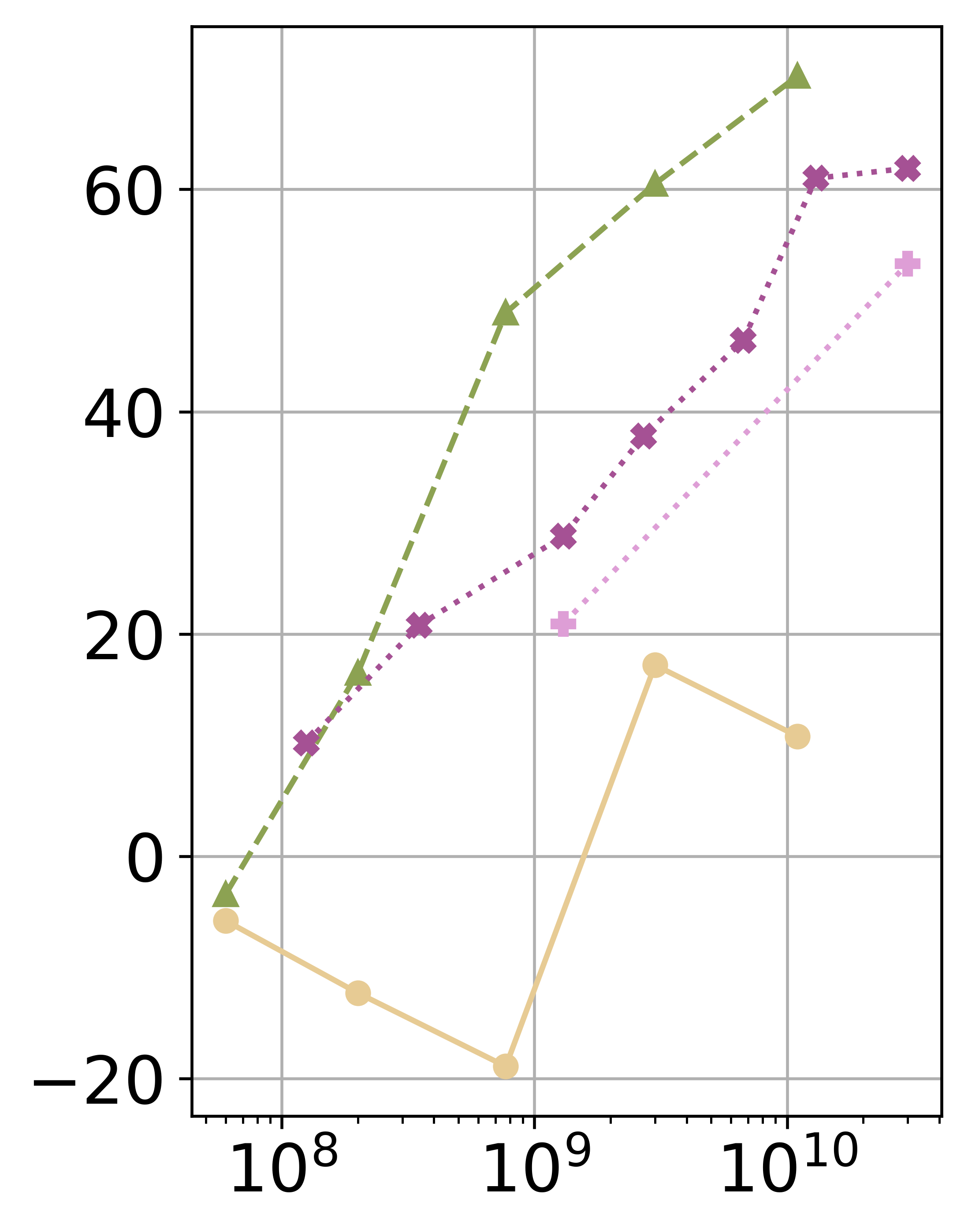}
        \caption{QA template}
    \end{subfigure}     
    \begin{subfigure}[b]{0.47\columnwidth}
        \centering
        \includegraphics[width=\columnwidth]{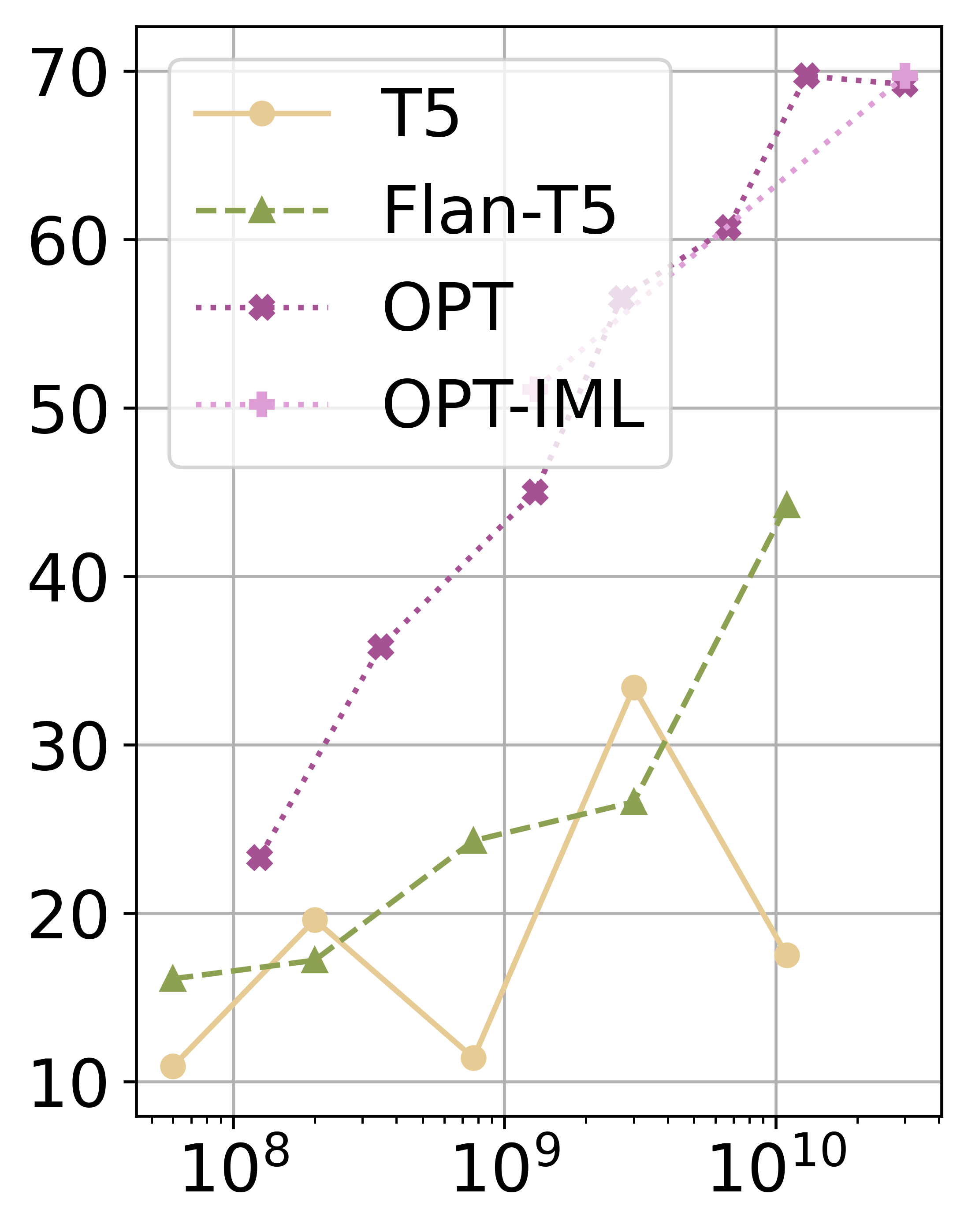}
        \caption{LC template}
    \end{subfigure}
    \caption{Spearman's rank correlation for the \emph{known for} relation type along with the model size.}
    \label{fig:correlation-with-model-size-know}
\end{figure}

\begin{figure}[!t]
    \centering
    \begin{subfigure}[b]{0.47\columnwidth}
        \centering
        \includegraphics[width=\columnwidth]{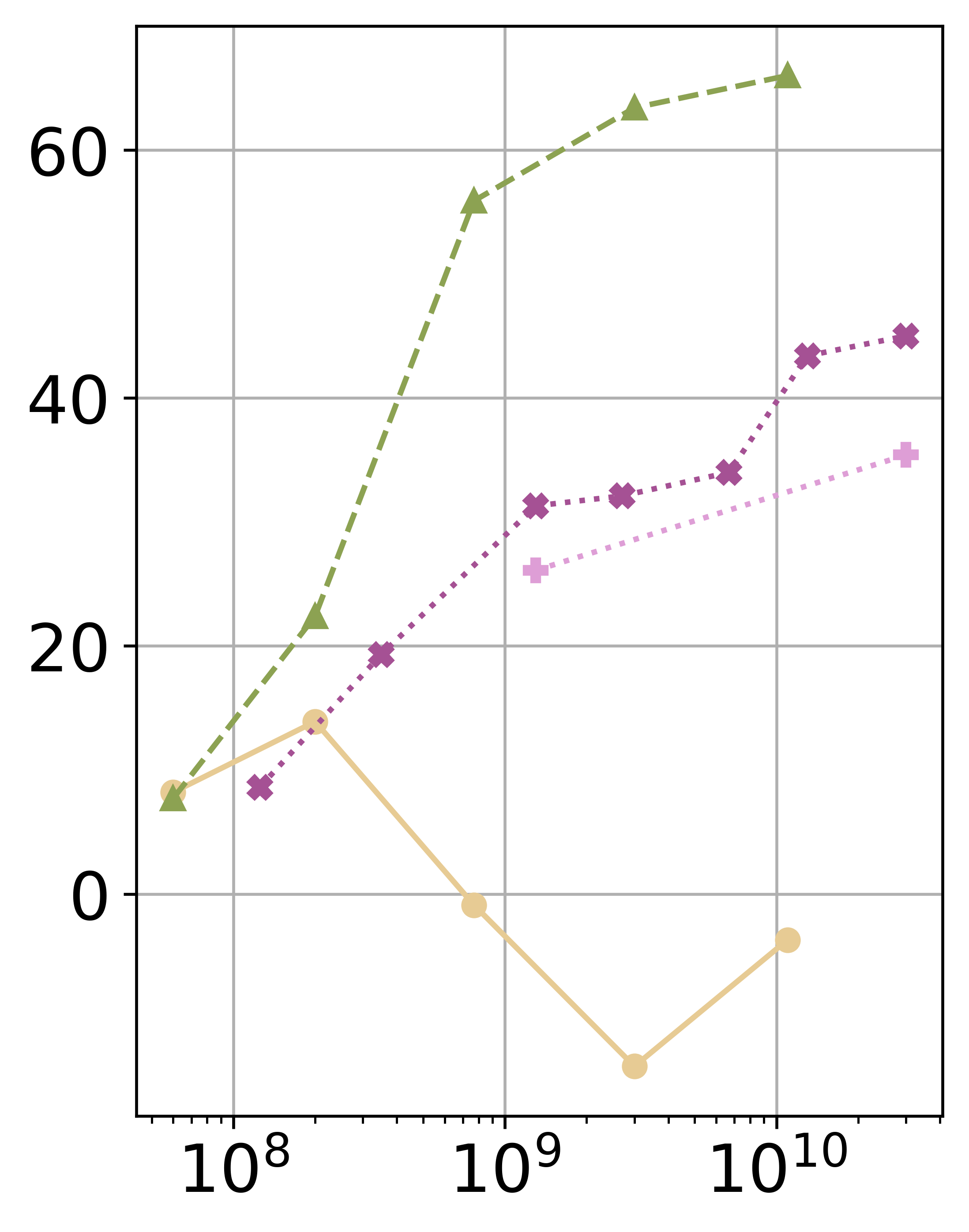}
        \caption{QA template}
    \end{subfigure}     
    \begin{subfigure}[b]{0.47\columnwidth}
        \centering
        \includegraphics[width=\columnwidth]{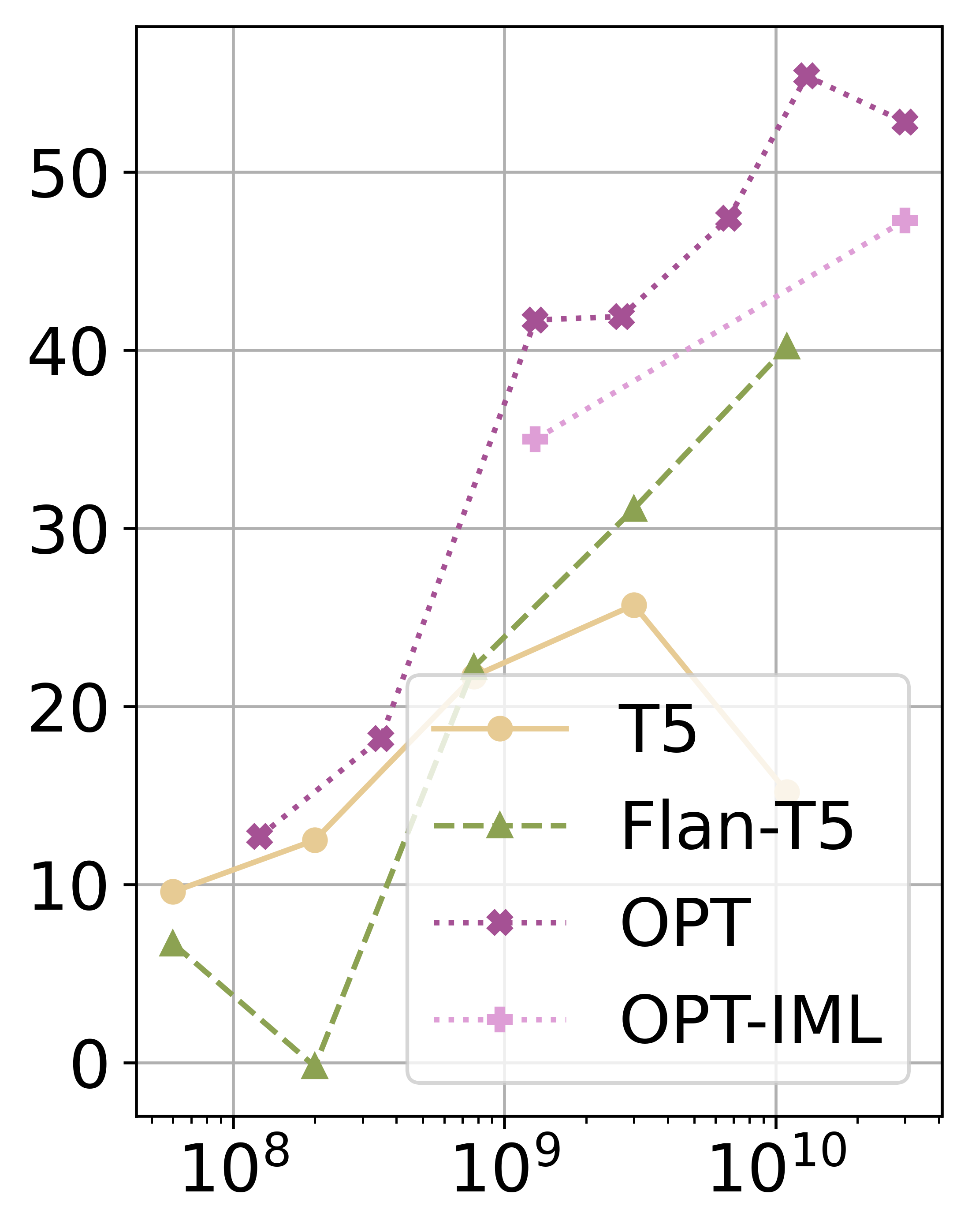}
        \caption{LC template}
    \end{subfigure}
    \caption{Spearman's rank correlation for the \emph{similar to} relation type along with the model size.}
    \label{fig:correlation-with-model-size-sim}
\end{figure}

\begin{figure}[!t]
    \centering
    \begin{subfigure}[b]{0.47\columnwidth}
        \centering
        \includegraphics[width=\columnwidth]{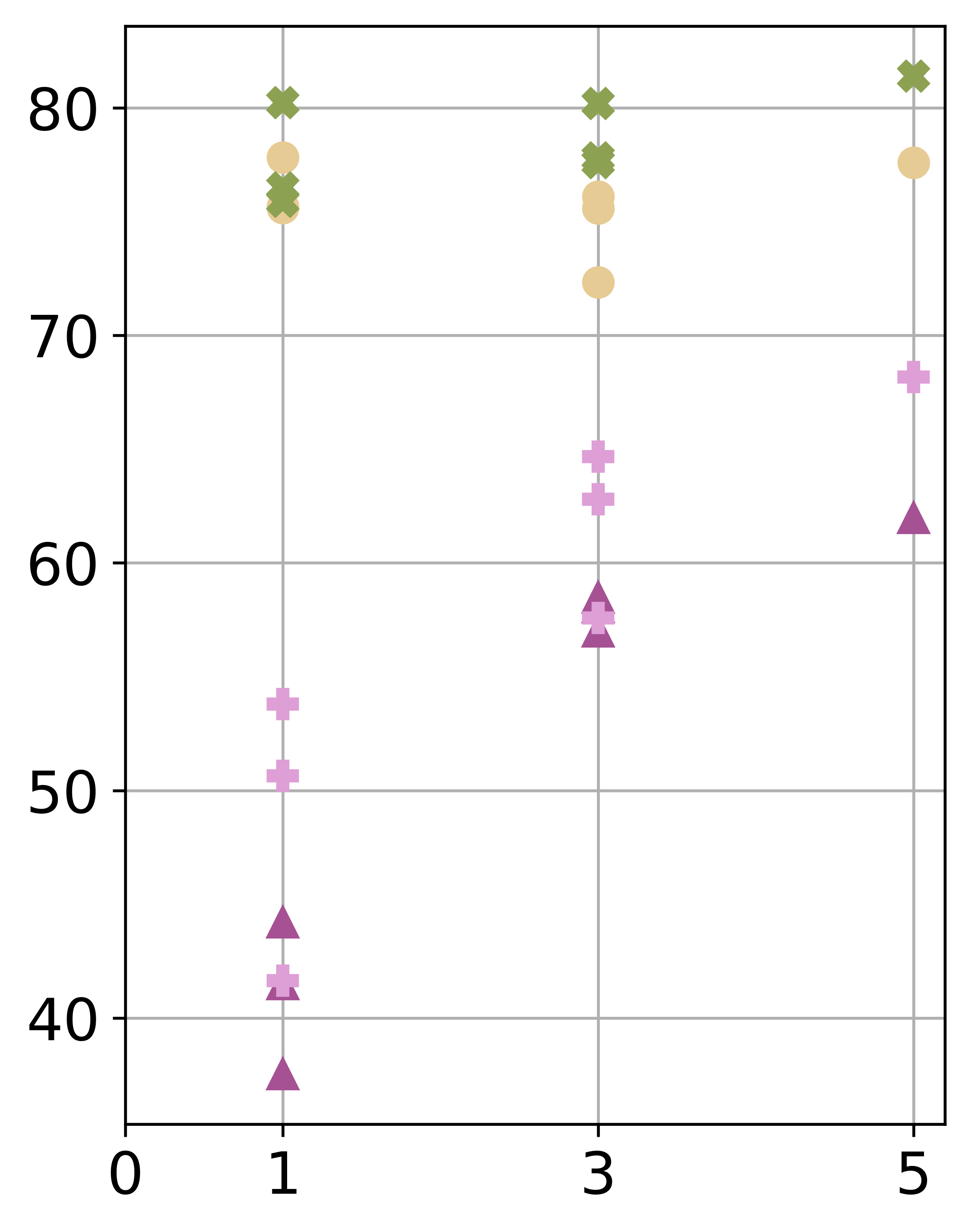}
        \caption{QA template}
        \label{fig:correlation-fewshot-ally-qa}
    \end{subfigure}     
    \begin{subfigure}[b]{0.47\columnwidth}
        \centering
        \includegraphics[width=\columnwidth]{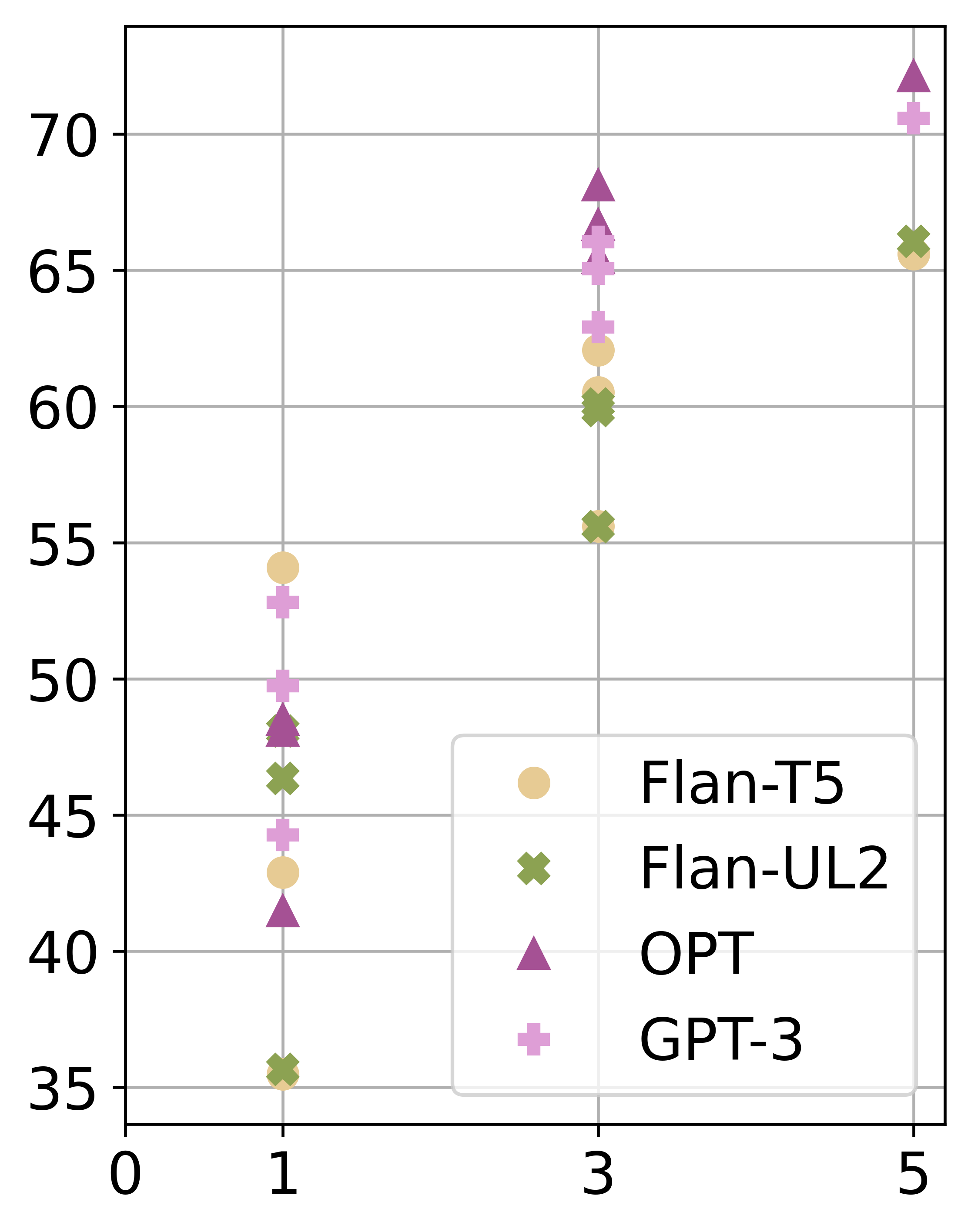}
        \caption{LC template}
        \label{fig:correlation-fewshot-ally-lc}
    \end{subfigure}
    \caption{Spearman's rank correlation for \emph{competitor/rival of} relation with different number of the prototypical examples.}
    \label{fig:correlation-fewshot-rival}
\end{figure}

\begin{figure}[!t]
    \centering
    \begin{subfigure}[b]{0.47\columnwidth}
        \centering
        \includegraphics[width=\columnwidth]{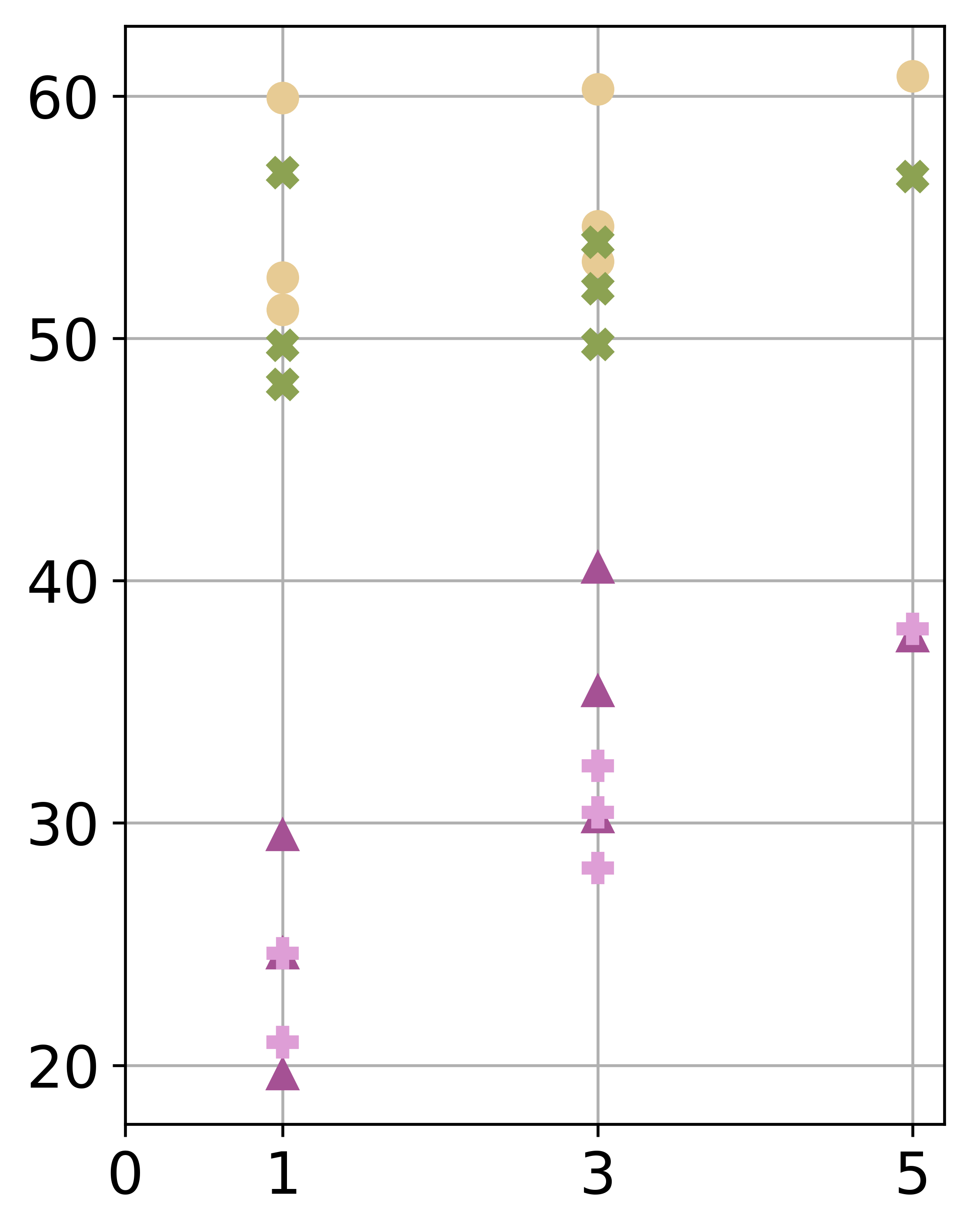}
        \caption{QA template}
        \label{fig:correlation-fewshot-ally-qa}
    \end{subfigure}     
    \begin{subfigure}[b]{0.47\columnwidth}
        \centering
        \includegraphics[width=\columnwidth]{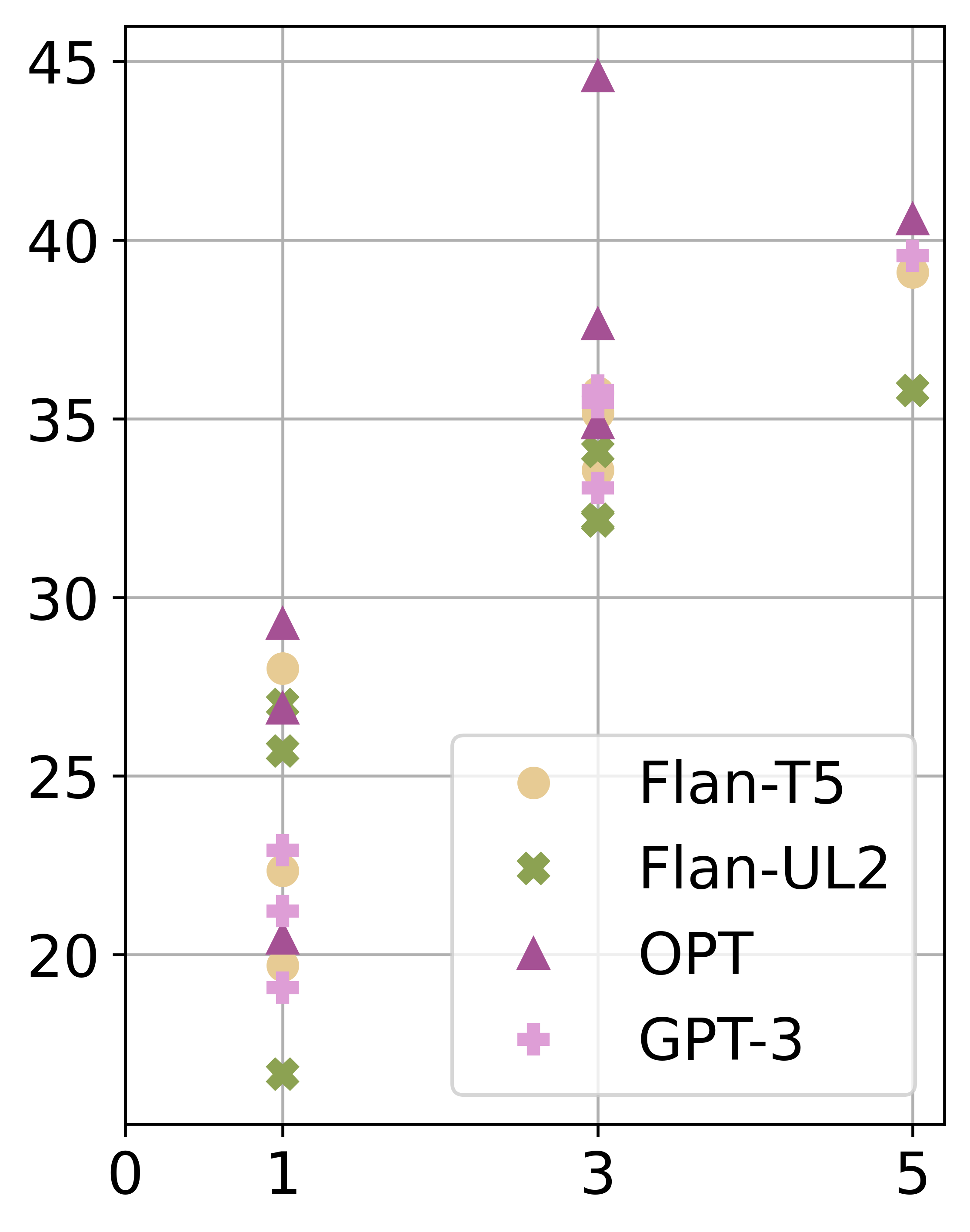}
        \caption{LC template}
        \label{fig:correlation-fewshot-ally-lc}
    \end{subfigure}
    \caption{Spearman's rank correlation for \emph{friend/ally of} relation with different number of the prototypical examples.}
    \label{fig:correlation-fewshot-ally}
\end{figure}

\begin{figure}[!t]
    \centering
    \begin{subfigure}[b]{0.47\columnwidth}
        \centering
        \includegraphics[width=\columnwidth]{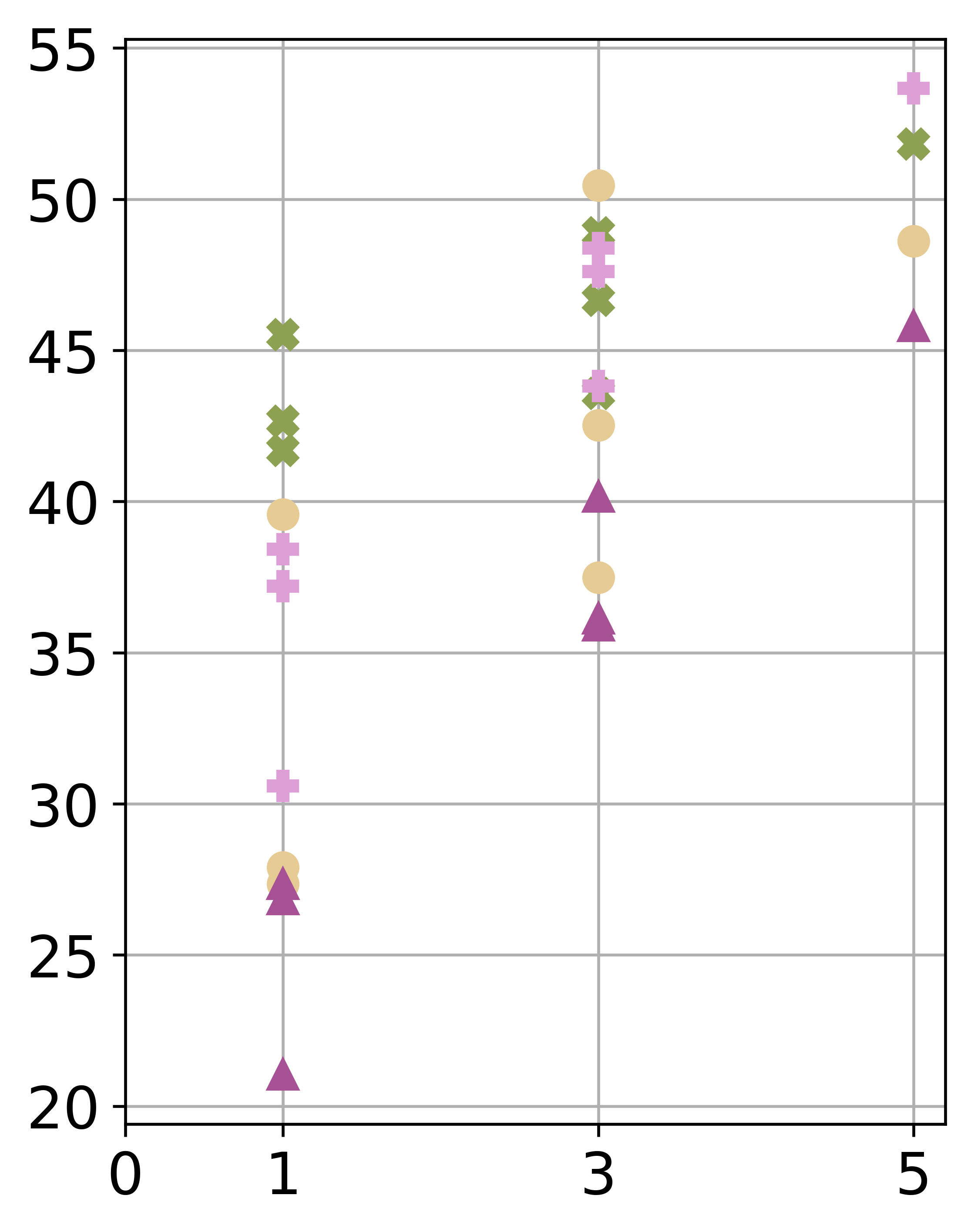}
        \caption{QA template}
        \label{fig:correlation-fewshot-inf-qa}
    \end{subfigure}     
    \begin{subfigure}[b]{0.47\columnwidth}
        \centering
        \includegraphics[width=\columnwidth]{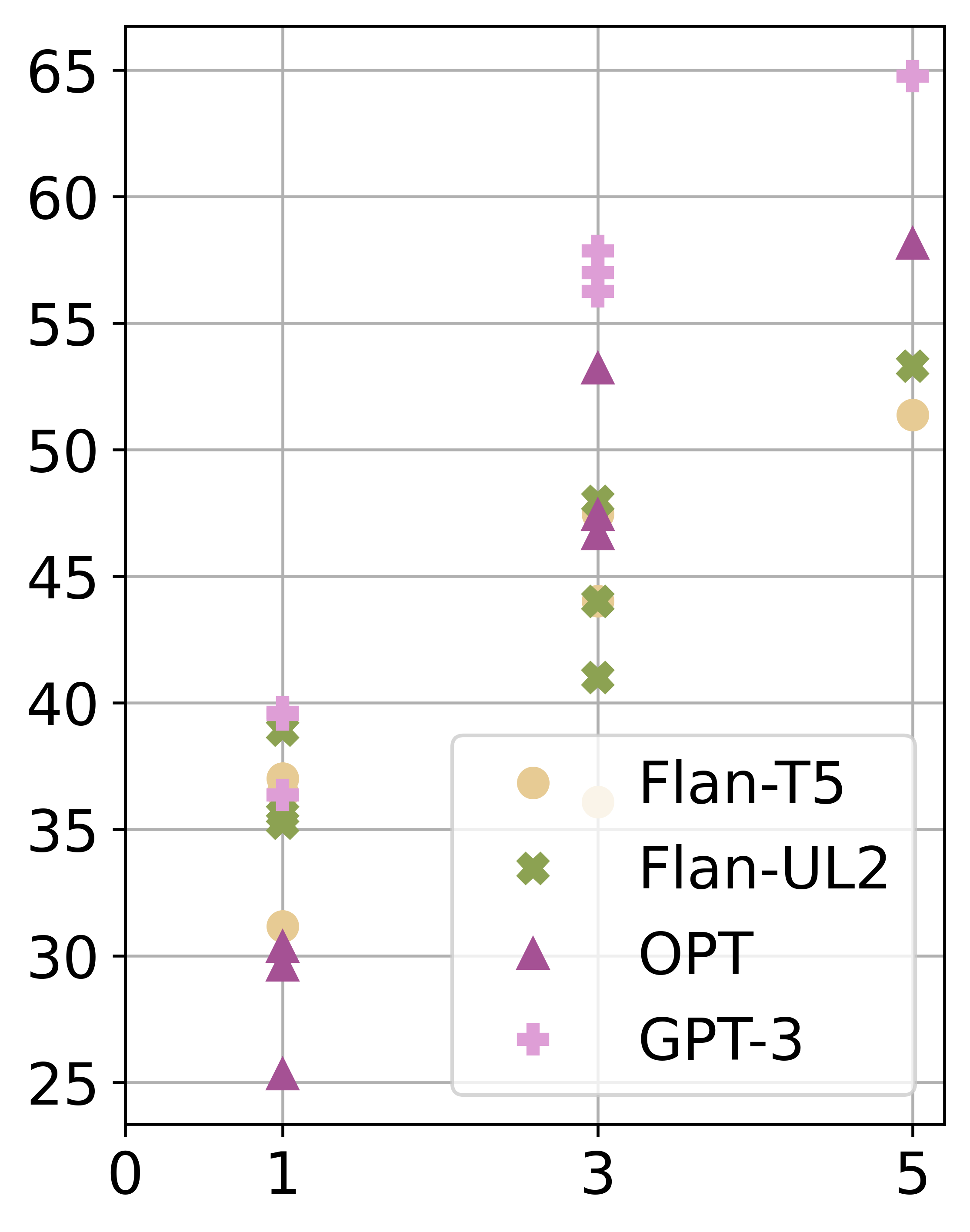}
        \caption{LC template}
        \label{fig:correlation-fewshot-inf-lc}
    \end{subfigure}
    \caption{Spearman's rank correlation for \emph{influenced by} relation with different number of the prototypical examples.}
    \label{fig:correlation-fewshot-inf}
\end{figure}

\begin{figure}[!t]
    \centering
    \begin{subfigure}[b]{0.47\columnwidth}
        \centering
        \includegraphics[width=\columnwidth]{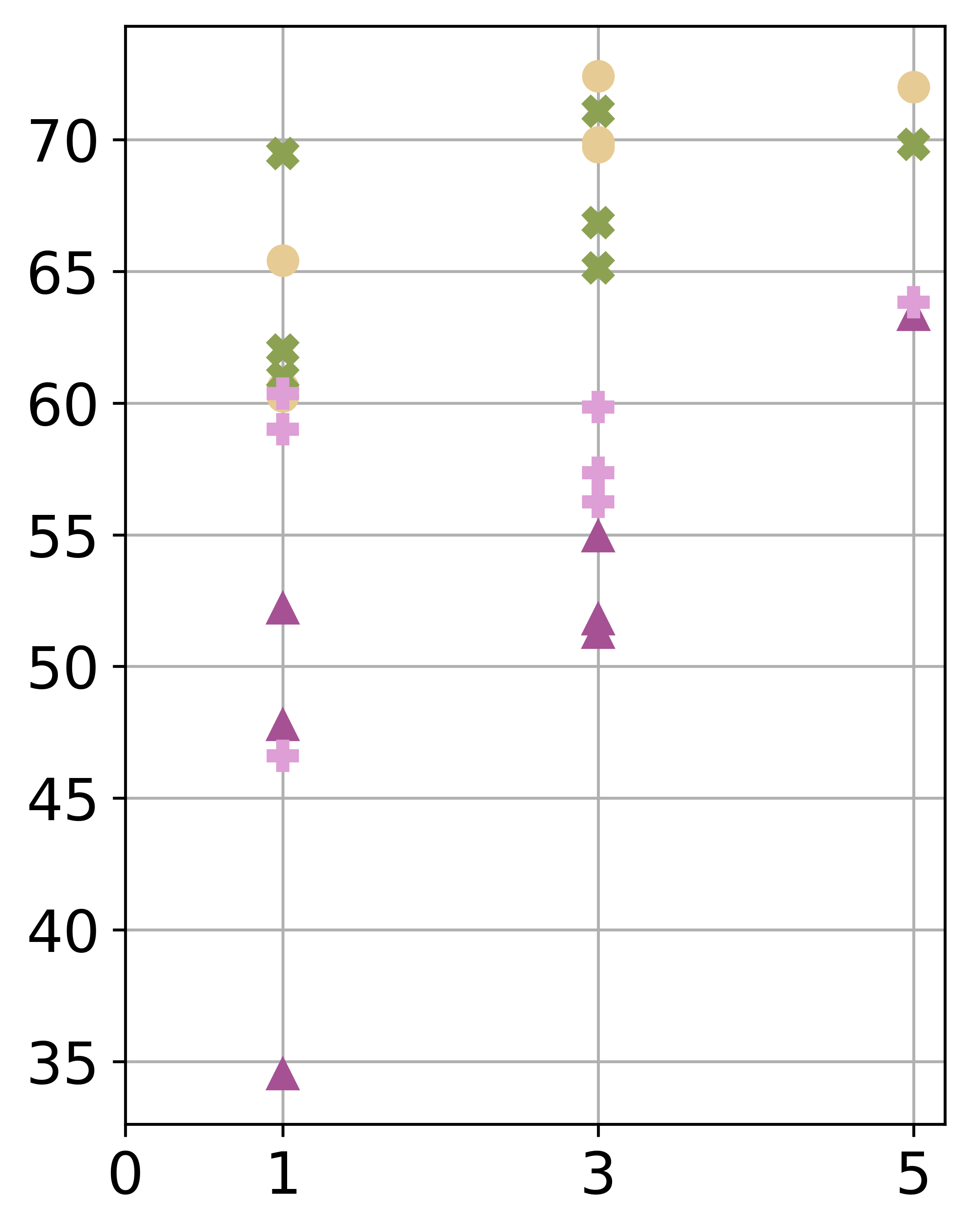}
        \caption{QA template}
        \label{fig:correlation-fewshot-know-qa}
    \end{subfigure}     
    \begin{subfigure}[b]{0.47\columnwidth}
        \centering
        \includegraphics[width=\columnwidth]{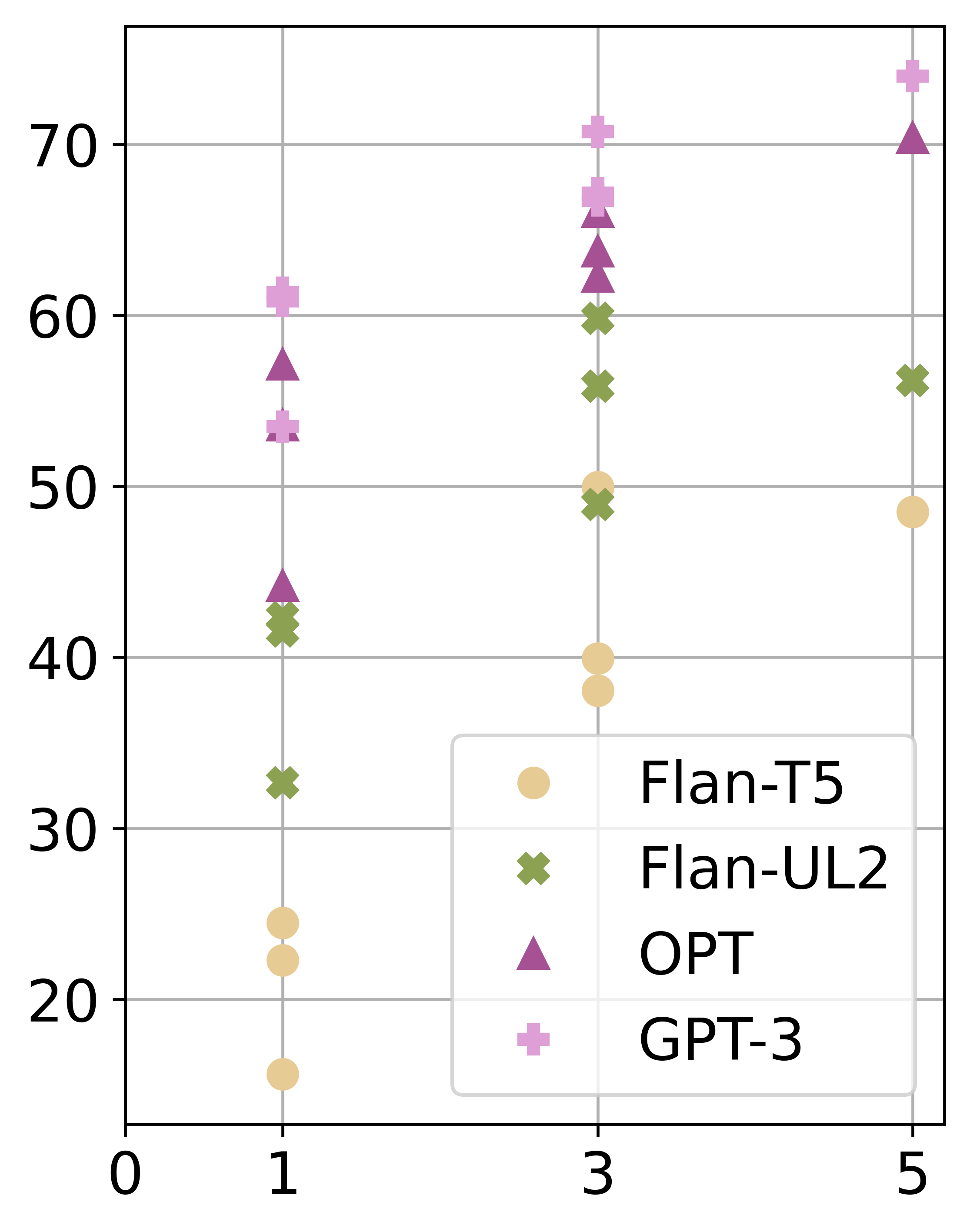}
        \caption{LC template}
        \label{fig:correlation-fewshot-know-lc}
    \end{subfigure}
    \caption{Spearman's rank correlation for \emph{known for} relation with different number of the prototypical examples.}
    \label{fig:correlation-fewshot-know}
\end{figure}

\begin{figure}[!t]
    \centering
    \begin{subfigure}[b]{0.47\columnwidth}
        \centering
        \includegraphics[width=\columnwidth]{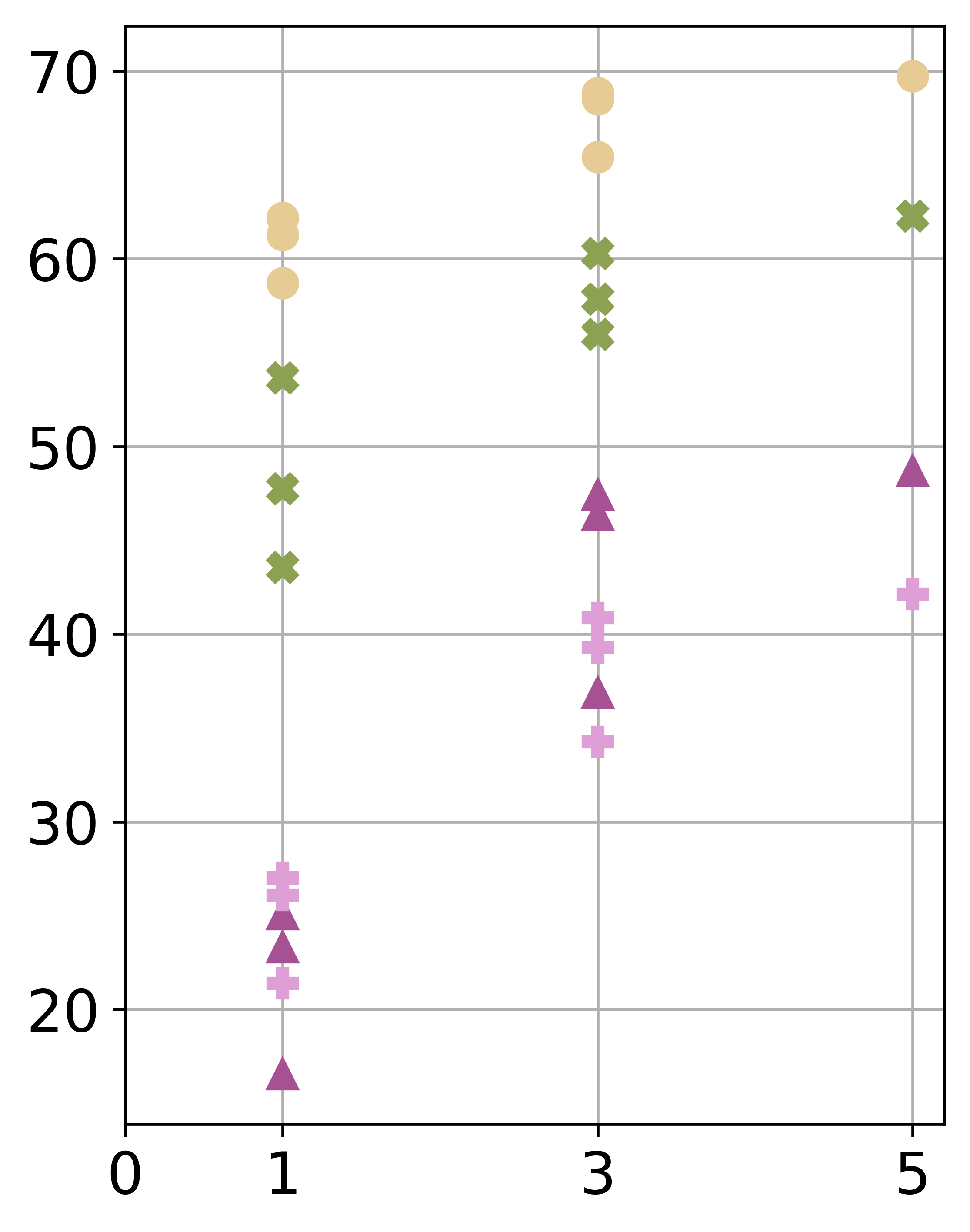}
        \caption{QA template}
        \label{fig:correlation-fewshot-sim-qa}
    \end{subfigure}     
    \begin{subfigure}[b]{0.47\columnwidth}
        \centering
        \includegraphics[width=\columnwidth]{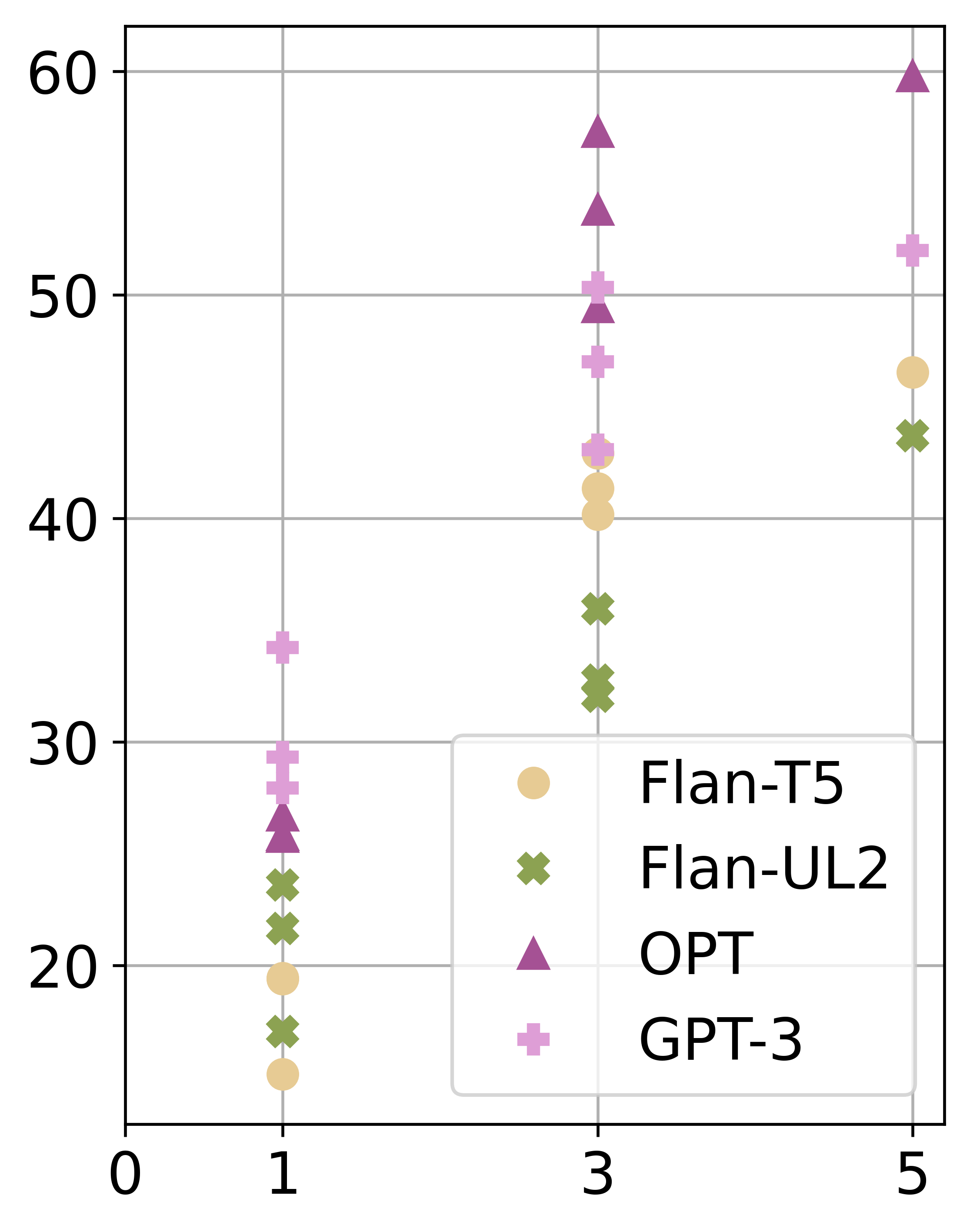}
        \caption{LC template}
        \label{fig:correlation-fewshot-sim-lc}
    \end{subfigure}
    \caption{Spearman's rank correlation for \emph{similar to} relation with different number of the prototypical examples.}
    \label{fig:correlation-fewshot-sim}
\end{figure}

\autoref{fig:correlation-with-model-size-rival},
\autoref{fig:correlation-with-model-size-ally},
\autoref{fig:correlation-with-model-size-inf},
\autoref{fig:correlation-with-model-size-know}, and
\autoref{fig:correlation-with-model-size-sim} show the performance improvement along with the model size for individual relation types.
\autoref{fig:correlation-fewshot-rival},
\autoref{fig:correlation-fewshot-ally},
\autoref{fig:correlation-fewshot-inf},
\autoref{fig:correlation-fewshot-know}, and
\autoref{fig:correlation-fewshot-sim} show the zero-shot and few-shot evaluation result for individual relation types.




\end{document}